\documentclass[11pt,oneside,reqno]{amsart}
\pdfoutput=1

\usepackage[english]{babel}
\usepackage{multicol}
\usepackage{amsmath}
\usepackage{amssymb}
\usepackage{amsfonts}
\usepackage{mathrsfs}
\usepackage{graphicx}
\usepackage{amsthm}
\usepackage{fancyhdr}
\usepackage{newlfont}
\usepackage{indentfirst}
\usepackage{geometry}
\usepackage{enumerate}
\usepackage{xcolor}
\usepackage[font={small,it}]{caption}
\usepackage[authoryear]{natbib}
\usepackage[colorlinks=true,linkcolor=blue,citecolor=blue]{hyperref}
\usepackage{placeins}
\usepackage{subfig}
\usepackage{color}
\usepackage{multirow}
\usepackage{booktabs}

\makeatletter
\newcommand\notsotiny{\@setfontsize\notsotiny{8}{7.5}}
\makeatother

\renewcommand{\baselinestretch}{1.5} 
\begin{document}

 \title[KerCNNs]{KerCNNs: biologically inspired lateral connections for classification of corrupted images}
 \author[N. Montobbio]{Noemi Montobbio$^{1,2}$}
 \thanks{$^1$ Dipartimento di Matematica, Universit\`{a} di Bologna, Italy.} \thanks{$^2$ CAMS Center of Mathematical Analysis, CNRS-EHESS, Paris, France.}
 \author[L. Bonnasse-Gahot]{Laurent Bonnasse-Gahot$^2$}
 \author[G. Citti]{Giovanna Citti$^1$}
 \author[A. Sarti]{Alessandro Sarti$^2$}
 \date{}
 \maketitle
 
 {\bf Keywords:} CNNs, Image classification, Visual cortex, Correlation kernels, Neurogeometry
 
 \begin{abstract}
  The state of the art in many computer vision tasks is represented by Convolutional Neural Networks (CNNs). Although their hierarchical organization and local feature extraction are inspired by the structure of primate visual systems, the lack of lateral connections in such architectures critically distinguishes their analysis from biological object processing. The idea of enriching CNNs with recurrent lateral connections of convolutional type has been put into practice in recent years, in the form of learned recurrent kernels with no geometrical constraints. In the present work, we introduce biologically plausible lateral kernels encoding a notion of correlation between the feedforward filters of a CNN: at each layer, the associated kernel acts as a transition kernel on the space of activations. The lateral kernels are defined in terms of the filters, thus providing a parameter-free approach to assess the geometry of horizontal connections based on the feedforward structure. We then test this new architecture, which we call KerCNN, on a generalization task related to global shape analysis and pattern completion: once trained for performing basic image classification, the network is evaluated on corrupted testing images. The image perturbations examined are designed to undermine the recognition of the images via local features, thus requiring an integration of context information -- which in biological vision is critically linked to lateral connectivity. Our KerCNNs turn out to be far more stable than CNNs and recurrent CNNs to such degradations, thus validating this biologically inspired approach to reinforce object recognition under challenging conditions.
 \end{abstract}

 \section{Introduction}
 
  Convolutional Neural Networks (CNNs) are a powerful tool that provides outstanding performances on image classification tasks. Major advances have been made since their introduction in the 1980s \citep{fuku}, thanks to the availability of large-scale datasets, as well as efficient GPU implementations and new regularization schemes. A notable example in this respect is the huge improvement of the state of the art performance reached by \citet{krizhevsky} on the ImageNet 2012 classification benchmark. However, there is still little insight into how the learning process of these algorithms develops and how the features of the data are encoded in the network structure. Some visualization techniques have been developed, such as the ``deconvolution''-based projection of activations onto the pixel space proposed by \citet{zeiler}, to identify the input stimuli that excite each feature map at a given layer of the network. Although this may provide some intuition on internal operations and simplify the diagnosis of the limitations of the models, there is still much to be understood about how exactly image information is coded in CNNs, and notably about how their functioning is related to human object processing. Indeed, although CNN models were initially inspired \citep{fuku,lecun_cnns} by the hierarchical model of the visual system of \citet{HW}, they display critical discrepancies w.r.t. biological vision in both structure and feature analysis.\\
  In \citet{baker} the authors show that, unlike in human vision, global shapes have surprisingly little impact on the classification output of the net: that is, CNNs turn out to learn mostly from local features. As such, CNN architectures are very unstable to small local perturbations, even when the global structure of the image is preserved and its content is still easily recognizable by a human observer. Along the same lines, it has been recently shown by \citet{brendel} that a very good classification accuracy on the ImageNet dataset can be reached through a model that only relies on the occurrences of local features, with no information on their spatial location in the image.\\
  Another key strand in unraveling the shortcomings in the internal processing of CNNs is the one related to ``adversarial attacks'': it has been shown \citep{szegedy} that a CNN can be caused to completely misclassify an image when it is perturbed in a way imperceptible to humans, but specifically designed to maximize the prediction error. A similar study is presented in \citet{fooled}, where the authors produce images that are unrecognizable to the human eye, but get labeled as one specific object class with high confidence by state-of-the-art CNNs.\\
  Besides, although the overall convolutional architecture has much in common with the process of feature extraction carried out in the visual pathways, its structure implements a \emph{purely feedforward} mechanism. On the contrary, the human visual system is well known to rely on both \emph{lateral} (intra-layer) and \emph{feedback} (top-down) recurrent connections for processes that are critical for object recognition, such as contour integration or figure-ground segregation \citep{gilbert,gromin,neumin,layton}. In recent years, several models have been proposed in which CNN architectures were enriched with some recurrent mechanism inspired by biological visual systems. In \citet{tang}, pre-trained feedforward models were augmented with a Hopfield-like recurrent mechanism acting on the activation of the \emph{last} layer, to improve their performance in pattern completion: partially visible objects converge to fixed attractor points dictated by the original whole objects. \citet{liang} introduced a ``Recurrent CNN'' architecture, where lateral connections of convolutional type are inserted in a regular feedforward CNN. A systematic analysis of the effect of adding lateral and/or feedback connections has been carried out by \citet{spoerer}, where the resulting architectures are trained and tested on a task of classification of cluttered digits. In Recurrent CNNs, lateral connections are learned, and no geometrical prior (apart from the ones given by the convolutional structure) is inserted. As such, these connections are determined by additional parameters that are completely independent of the feedforward architecture.\\
  
  In this work, we propose to modify the classical CNN architecture by inserting lateral connections defined by \emph{structured} kernels, containing precise geometric information specific of each layer. The new architecture will be referred to as KerCNN. The kernel associated to a convolutional layer implements a measure of \emph{correlation} between neurons of that layer, inspired by the connectivity model of \citet{neuro,metric}, and acts as a transition kernel on the corresponding activation. As will be discussed in Section \ref{archi}, the lateral contribution is defined by an iterative update rule similar to the recurrent mechanism of \citet{liang}, although carefully modified to implement a biologically plausible propagation of neural activity. Most importantly, the lateral kernels themselves are not learned, but rather they are constructed to allow diffusion in the metric defined by the learned filters. In particular, they establish a link between the geometrical properties of feedforward connections and horizontal connectivity, being defined as a function of the convolutional filters. This also implies that such kernels do not depend on any additional trainable parameters: therefore, their insertion does not increase the original network's complexity in terms of number of parameters, which allows a fair comparison in performance.\\
  The main point that we wish to make is that the insertion of these connections allows the networks to \emph{spontaneously} implement perceptual mechanisms of global shape analysis and completion. Therefore, we shall examine the ability of the models to \emph{generalize} an image classification task to data corrupted by a variety of different perturbations: these include occlusions \citep[as in][]{tang}, local contour disruption \citep[as in][]{baker} and adversarial attacks via the Fast Gradient Sign Method (FGSM) of \citet{fgsm}. We stress that the data perturbations are only inserted in the \emph{testing} phase -- that is, the models are not optimized to classify corrupted images.\\
  We will fix a base 2-layer CNN model and modify it by inserting our structured lateral connections in one or both layers. We will then compare the performance of the base CNN with the one of the different KerCNN models, obtained by varying the number of iterations of the update rule for each layer. We first present an extensive analysis of our results for the classical MNIST dataset \citep{mnist}. As will be shown in Section \ref{resmnist}, KerCNNs turn out to improve the base CNN's classification accuracy on degraded images by up to $\sim25$ points, while preserving the same performance on the original (not corrupted) testing images. We also compare the KerCNN models with the ``RecCNN'' obtained by adding recurrent connections to the base model as in \citet{spoerer} -- where the number of parameters of the networks is matched by decreasing the size of feedforward filters, to compensate for the additional recurrent parameters. In particular, for each task we inspect the performance of the \emph{best} KerCNN and RecCNN architectures (i.e. the ones with the optimal number of iterations), and our results show that our biologically inspired model outperforms the recurrent one in practically all experiments, see Section \ref{cfr}. We will conclude the paper by giving a synthetic account on the same study carried out on different datasets, namely Kuzushiji-MNIST \citep{kmnist}, Fashion-MNIST \citep{fashion} and CIFAR-10 \citep{cifar}. The choice of the two MNIST-like datasets was driven by their being very homogeneous, allowing for a meaningful interpretation of our results in terms of the characterizing features of the images. On the other hand, our tests on CIFAR-10 show that this technique can be extended to richer datasets, and notably to natural images.\\
  A noteworthy feature of our model is that it somewhat links two approaches to image treatment that are classically seen as opposites, namely ``geometrical'' methods and ``data-driven'' methods. The former rely on \emph{a priori} assumptions based on mathematical modeling either of the structure of the data or of the task, e.g. variational techniques for inpainting \citep{ambmasn,inpainting}; the latter instead are designed to \emph{learn} patterns and convenient representations from the statistics of a dataset through optimization of a loss function related to the task. In KerCNNs these two aspects coexist, since the metric structure that we define on each layer of the network is directly induced by the learned filters.

 \section{Preliminaries}
 
 In this section, we shall give an overview on some biological notions and computational methods that will be of interest throughout the paper. CNNs are a particular kind of deep neural network architecture, designed in analogy with information processing in biological visual systems \citep{fuku,lecun_cnns}. In addition to the hierarchical organization typical of deep architectures, translation invariance is enforced in CNNs by local convolutional windows shifting over the spatial domain. This structure was inspired by the localized receptive profiles of neurons in the early visual areas, and by the approximate translation invariance in their tuning. Although the analogy with biological vision is strong, the feedforward mechanism implemented in CNNs is a simplified one, and it does not take into account \emph{all} of the processes contributing towards the interpretation of a visual scene. In the following, we first describe some structures of the visual pathways with a focus on the primary visual cortex (V1), and review some mathematical models of vision; we then recall the main features of feedforward convolutional architectures, and finally outline the Recurrent CNN models of \citet{liang} and \citet{spoerer}.
 
 \subsection{Feedforward and lateral connectivity of V1}\label{v1}
 
 The primary visual cortex (V1) implements the first stage of cortical processing of a visual stimulus. It receives the retinal signal after a first \emph{subcortical} processing stage, and it sends information to ``higher'' cortical areas performing further processing. These junctions form a connectivity of \emph{feedforward} type, since they link the zones of the visual pathways in a sequential way, generating a hierarchy starting from the retina.\\
 Through the above-mentioned connections, each visual neuron is linked to a specific domain $D$ of the retina which is referred to as its \emph{receptive field} (RF). The reaction of a cell to a punctual luminous stimulation applied at a point $(x,y) \in D$ can be of excitatory or inhibitory type, with different modulation: this can be described by a function $\psi : D \rightarrow \mathbb{R}$, called the \emph{receptive profile} (RP) of the cell, whose values are positive when the cell is excited and negative when it is inhibited. The RPs of certain types of visual neurons are shown to act, at least to a first approximation, as \emph{linear filters} on the optic signal. This means that the response of the cell to a visual stimulus $I$, defined as a function on the retina, is given by the integral of $I$ against the profile $\psi$ of the neuron, computed over its receptive field $D$:
\begin{equation}\label{linfilt}h := \int_D I(x,y)\psi(x,y)dxdy.\end{equation}
 In these cases, the shape of the RP contains information about the \emph{features} that it extracts from a visual signal. For example, the local support\footnote{The \emph{support} of a function $\psi$ is defined as the closure of the set $\{x \: : \: \psi(x)\neq 0\}$.} of $\psi$ makes it sensitive to \emph{position}, i.e. the neuron only responds to stimuli in a localized region of the image. Or again, a receptive profile with an elongated shape will be sensitive to a certain \emph{orientation}, i.e. it will respond strongly to stimuli consisting of bars aligned with this shape. This is the case for \emph{simple cells}, a class of neurons of V1 showing orientation selectivity due to their strongly anisotropic RPs, first discovered by \citet{HW}.\\
 
 The processing performed by the human visual system allows to efficiently group local items into extended contours, and to segregate a path of elements from its background. This implies that the perception of a local edge element in a visual stimulus is influenced by the perception of the surrounding oriented components: this perceptual phenomenon has been described through the concept of \emph{association field} \citep{field}, characterizing the geometry of the mutual influences between local oriented elements in the perceived image, depending on their orientation and reciprocal position. These psychophysical experiments suggest that a local analysis is not sufficient to correctly interpret a visual scene: from the physiological point of view, this means that the activity of V1 neurons is not only influenced by the feedforward signal received from the preceding visual areas, but also by \emph{intracortical} connections with surrounding V1 cells. In fact, the reciprocal influences described by association fields are thought to be neurally implemented in V1 through a kind of long-range connections referred to as \emph{lateral} (or \emph{horizontal}), whose orientation specificity and spatial extent is compatible with association fields. Indeed, V1 horizontal connections show facilitatory influences for cells that are similarly oriented; moreover, the connections departing from each neuron spread anisotropically, concentrating along the axis of its preferred orientation \cite[see e.g.][]{bosking}.\\
 
 The functional architecture of V1 and the related perceptual phenomena have been described in a variety of mathematical models. The set of RPs of simple cells is typically represented by a bank of linear filters $\{\psi_p\}_{p \in \mathcal{G}} \subseteq L^2(\mathbb{R}^2)$, where $\mathcal{G}$ is a set of indices parameterizing the family. Each $p \in \mathcal{G}$ can be thought of as representing the features extracted by the filter $\psi_p$: in these terms, we can refer to $\mathcal{G}$ as the \emph{feature space} associated to the bank of filters $\{\psi_p\}_{p \in \mathcal{G}}$. This set often has the product form $\mathcal{G}=\mathbb{R}^2\times\mathcal{F}$, where the parameters $(x,y) \in\mathbb{R}^2$ determine the retinal location of the RF, while $f\in\mathcal{F}$ represents the other local image features extracted by each filter. This is the case for the work of \citet{bresscow03}, where each V1 cell is labeled by a spatial index and a ``feature index'', and the evolution in time $t\mapsto a(p,t)$ of the activity of the neural population at $p \in \mathbb{R}^2\times \mathcal{F}$ is assumed to satisfy a Wilson-Cowan equation \citep{wilcow}:
 \begin{equation}\label{meanfield}
  \partial_t a(p,t) = -\alpha \: a(p,t) + s\left(\int \phi(p,p')a(p',t)dp' + h(p,t)\right).
 \end{equation}
 Here, $s$ is a nonlinear activation function; $\alpha$ is a decay rate; $h$ is the feedforward input corresponding to the response of the simple cells in presence of a visual stimulus, as in (\ref{linfilt}); and the kernel $\phi$ weights the strength of horizontal connections between $p$ and $p'$. A possible way to obtain a measure of this connectivity is by means of differential geometry tools. A breakthrough idea in this direction has been that of viewing the feature space $\mathcal{G}=\mathbb{R}^2\times\mathcal{F}$ as a fiber bundle with basis $\mathbb{R}^2$ and fiber $\mathcal{F}$. This approach first appeared in the works of \citet{koenderink} and \citet{hoffman}. It was then further developed by \citet{petitond} and \citet{cs06}. In the latter work, the model is written in the Lie group $\mathbb{R}^2 \times S^1$ by requiring the invariance under roto-translations: here, the feature index explicitly represents a local orientation $\theta$. More generally, it can also contain information about other variables such as scale, curvature or even velocity \cite[see e.g.][]{symplectic,abbfav,bccs}. Another strand of research is linked to statistics of natural images \citep[see e.g.][]{augzuck,kruger,sigman,edge-stat}. In \citet{edge-stat}, the statistics of edge co-occurrence in natural images are fitted to a Fokker-Planck kernel in $\mathbb{R}^2 \times S^1$; such kernel has been proposed as a connectivity weight $\phi$ to insert in (\ref{meanfield}) by \citet{perceptual}. 
 
 \subsubsection{A kernel model for lateral connectivity}\label{neuroker}
 
 A different connectivity kernel, induced by a structure of metric space associated to the RPs of simple cells, has been introduced in \citet{neuro}. The core of the model is the definition of a kernel describing the interactions between local elements, which determines a metric structure directly induced by the shape of the RPs of V1 simple cells. This local correlation kernel is then propagated through an iterative procedure, to yield a wider kernel modeling long-range connections.
 
 The starting point is a family of filters $\{\psi_p\}_{p\:\in\:\mathcal{G}} \subseteq L^2(\mathbb{R}^2)$ modeling the set of V1 simple cells. The \emph{local} connectivity of V1 is represented by the following \emph{generating kernel} on $\mathcal{G} \times \mathcal{G}$: 
 \begin{equation}\label{defK}
 K(p,p_0) := Re\langle\psi_p,\psi_{p_0}\rangle_{L^2} = Re \left(\int_{\mathbb{R}^2} \psi_p(x,y) \: \overline{\psi_{p_0}(x,y)} \: dx \: dy \right) \quad \forall p,p_0 \in \mathcal{G}.
 \end{equation}
 The kernel $K$ is constructed to provide a measure of \emph{correlation} between RPs. In fact, if the filters are normalized to have squared $L^2$-norm equal to some number $\eta>0$, then 
 \[d(p,p_0) := \| \psi_p - \psi_{p_0} \|^2_{L^2} = 2\big(\eta - K(p,p_0)\big).\]
 This means that $K$ expresses the correlation w.r.t. the $L^2$ distance between the filters. Note that this also defines a metric $d$ onto the feature space.
 
 The generating kernel has a local sense, since it only describes the reciprocal influences between simple cells with overlapping RFs. The action of $K$ is then iterated to model the long-range connectivity. Given a starting point $p_0$, the local kernel around it is first passed through a nonlinear activation function $\nu$ and a normalization operator
 $N$ \citep[see also][]{coiflaf}, thus defining:
 \begin{equation}\label{laf}K_1^{p_0}(p) := N[\nu(K)](p,p_0).\end{equation}
 The iterative procedure yielding the propagation is then given by
 \begin{equation}\label{iteration} K_{n}^{p_0}(p) := \int_\mathcal{G} N[\nu(K)](p,q) \: K_{n-1}^{p_0}(q) d\mu(q) 
 \quad \forall n>0.\end{equation}
 Here, $\mu$ is the spherical Hausdorff measure \citep{hausd} associated to the distance $d$ on $\mathcal{G}$.
 
 The geometrical structure encoded in this kernel is shown in \citet{neuro} to be compatible with the properties of V1 horizontal connections, and with the perceptual principles synthesized by association fields. Results are also shown for a bank of filters arising from an unsupervised learning algorithm: this shows that meaningful information of the geometry of horizontal connections can be recovered from numerically known filters, thus motivating the present work.\\
 
 We conclude this section with an important remark on the action of the correlation kernel of \cite{neuro} as an operator acting on functions defined on $\mathcal{G}$. Given a function 
 \[F_0 : \mathcal{G} \longrightarrow \mathbb{R},\]
 the action of the \emph{propagated} kernel onto $F$ can then be expressed by
 \begin{equation}\label{imevol} H_{n}[F](p_0) := \int_\mathcal{G} K_n^{p_0}(p) \: F_0(p) d\mu(p).\end{equation}
 Note that, by substituting Eq. (\ref{iteration}) into Eq. (\ref{imevol}), we get:
 \begin{align*}H_{n}[F](p_0) = 
 \underbrace{H_1 \ldots H_1}_n[F](p_0).\end{align*}
 This means that applying the $n$-th step kernel to $F_0$ is equivalent to applying $n$ times the local kernel to $F_0$. In the following, we will take as functions $F_0$ the \emph{activations} obtained by mapping a signal to a feature space. In the case of V1, this signal is a retinal image $I$ and the activation in presence of $I$ is a function of the cortical coordinates $p \in \mathcal{G}$:
 \[ F_0(p) := s\left(\int I(x,y)\psi_p(x,y) dxdy\right),\]
 where $s$ is a nonlinear activation function. Updating this activation through the connectivity kernel means taking into account the contextual influences in modeling the response of V1 to the image $I$.
 
 \subsection{CNNs for image classification}
 
 In order to fix the notations that will be used in later sections, we recall here the typical structure of a CNN, with a focus on image classification tasks. We refer to \citet{rawatwang} for an exhaustive review on this topic. As mentioned in Section \ref{v1}, the processing of early visual areas is classically modeled as the mapping of an image to a feature space through a bank of filters with a localized support. The first convolutional layer of a CNN implements an analogous mechanism, typically  defined as follows:
 \begin{equation}\label{lifting1}
  h_1(i,j,k) = s\left( \sum_{i',j',c} \psi^1_k(i',j',c)\:\cdot\:I(i-i',j-j',c)\; +\; b_1(k) \:\right),
 \end{equation}
 where $s$ is a nonlinear activation function. A popular choice for it in recent literature is the Rectified Linear Unit (ReLU) $s(z) := \max(0,z)$ \citep{relu,krizhevsky}. Here, the input image $I$ is an $H_0\times W_0\times n_0$ tensor, where $H_0$ and $W_0$ denote the height and width of the image in pixels, while $n_0$ is the number of \emph{channels}: $n_0=1$ if $I$ is a grayscale image, $n_0=3$ if it is RGB. The bank of filters $\{\psi^1_k\}_{k=1,\ldots, n_1}$ of the first layer is a $d_1 \times d_1 \times n_1 \times n_0$ tensor: $d_1 \times d_1$ is the \emph{spatial size} of the filters, $n_1$ is the number of filters and $n_0$ is the number of channels of each filter, matching the number of channels of the input. The convolution between $I$ and the filters $\psi^1_k$ gives an $H_1\times W_1 \times n_1$ tensor, to which a \emph{bias} vector $b_1 \in \mathbb{R}^{n_1}$ is added along the third component, to obtain the output $h_1$ of the layer. Note that the number of filters $n_1$ defines the number of channels of the output of the layer. Written in a more compact notation, Eq. (\ref{lifting1}) reads:
 \[h_1 = s(\psi^1 \ast I + b_1).\]
 The subsequent convolutional layers are defined similarly: for each $l \in \{1,\ldots,L\}$ we have a bank of filters $\{\psi^l_k\}_{k=1,\ldots, n_l}$ defined by a $d_l \times d_l \times n_l \times n_{l-1}$ tensor, and a bias vector $b_l$ of length $n_l$. The number of channels of the filters varies across layers according to the number of channels of the inputs they receive. In particular, since the output $h_{l-1}$ of the $(l-1)$-th layer has $n_{l-1}$ channels, each of the filters $\psi^l_k$ applied to it must have $n_{l-1}$ channels as well. The activation of the $l$-th layer in terms of the output $h_{l-1}$ of the preceding layer is given by
 \[h_l = s(\psi^l \ast h_{l-1} + b_l).\]
 Another layer that can optionally be interposed between convolutional layers consists of the application of a \emph{pooling} operator $\mathcal{P}$: this performs a downsampling of its input over the \emph{spatial variables} (i.e. the ``depth'' dimension remains unchanged), typically by taking the maximum or by averaging over small neighborhoods. For instance, if a pooling layer is applied to an activation $h_l$ of size $H_l\times W_l \times n_l$ over 2$\times$2 squares, then the output $\mathcal{P}(h_l)$ will be an $\frac{H_l}{2}\times \frac{W_l}{2} \times n_l$ tensor. This downsampling operation reduces the dimensionality and introduces invariance to small shifts and distortions. The insertion of pooling layers has a neural motivation as well: the receptive fields of visual neurons tend to get wider and wider moving towards higher cortical layers, and subsampling the spatial dimension of a feature space is equivalent to taking filters with a wider support in the next layer.\\
 The final layer of the network is typically \emph{fully connected}: the output $h_L$ of the last convolutional layer, which is a tensor of size $H_L \times W_L \times n_L$, is ``flattened'' to a vector $\tilde{h}_L$ of length $S = H_L \cdot W_L \cdot n_L$ and transformed as follows:
 \[\Psi\cdot\tilde{h}_L + b,\]
 where $\Psi$ is an $S \times n$ matrix of trainable weights and $b$ is a bias vector of length $n$. This yields a vector of length $n$ as output: in the case of multiclass classification, $n$ must be the number of classes. It is also not uncommon to have multiple fully connected layers, with nonlinear activation functions interposed between them -- in this case, only the length of the \emph{last} output vector needs to match the number of classes. A \emph{softmax} function $\rho$ is then typically applied to the final output vector:
 \[\rho(v)_i = \frac{e^{v_i}}{\sum_j e^{v_j}}.\]
 The softmax function gives a vector whose entries are real numbers between 0 and 1 that sum to 1: this can be interpreted as a vector of probabilities, where each entry represents the ``score'' of the corresponding class.\\
 The most common loss function for multiclass classification is the \emph{cross entropy} between the output $y = y(I)$ and the target vector $T = T(I)$ containing the ``true'' probabilities associated to the input $I$:
 \begin{equation}\label{loss} \mathcal{L}(y,T) = -\sum_{i=1}^{n} T_i \log(y_i) = - \log(y_{i(I)}),\end{equation}
 where $i(I)$ is the correct class for $I$. The last equality holds since $T_{i(I)} = 1$ and $T_i = 0$ for each $i\neq i(I)$.

 \subsection{Recurrent CNNs (RecCNNs)}\label{reccnns}
 
 The human visual system, as outlined in Section \ref{v1}, relies not only on a hierarchical transmission of signals, but also on a \emph{horizontal} spreading of information. On the other hand, the sequential structure of a CNN implements a purely feedforward mechanism: the output of each layer only depends on the activation of the preceding layer. Recurrent CNNs \citep{liang,spoerer}, referred to as RecCNNs in the following, are a modification of this kind of architecture, where \emph{lateral} connections of convolutional type are added to a regular CNN, yielding an equation analogous to (\ref{meanfield}). This means that the network includes not only connections from one layer to the next one, but also connections from a layer to itself, ruled by ``horizontal'' connectivity weights. As in (\ref{meanfield}), this is described through an evolution in time. The activation of the $l$-th hidden layer at time $t$, which we denote by $h_l^t$, is a function of:
 \begin{itemize}
  \item $h_{l-1}^t$ (the output of the preceding layer at the same time step $t$);
  \item $h_l^{t-1}$ (the output of the same layer at time $t-1$).
 \end{itemize}
 Specifically, following the notations introduced for CNNs, we have:
 \begin{equation}\label{rcnnrule} h_l^t = s\big( \underbrace{\phi^l \ast h_l^{t-1}}_{\text{lateral}} \:+\: \underbrace{\psi^l \ast h_{l-1}^t}_{\text{feedforward}} \:+\: b_l \big)\end{equation}
 for all $t,l>0$, where $h_0^t = I$ for all $t>0$, and $h_l^0 \equiv 0$ for all $l>0$. Here, $\phi^l$ denotes the bank of convolutional filters defining the lateral connections at the $l$-th layer. Note that their introduction results in an additional set of parameters in the architecture w.r.t. a standard CNN.\\
 Recurrent neural networks (RNNs) are often employed to process sequential inputs, e.g. audio recordings, video or text. In such cases, a new input $I^t = h_0^t$ is fed into the network at each time step. On the contrary, in RecCNNs the input image $I$ is \emph{static}, i.e. it is kept fixed at each time step: the time variable only affects the processing.

 \section{Kernel CNNs (KerCNNs)}
 
 The lateral connections in RecCNNs are completely learned and independent of the feedforward ones. As such, the inclusion of these connections in a CNN increases its complexity in terms of trainable parameters. We propose a different modification of a CNN, obtained by introducing convolutional lateral connections with kernels constructed according to the connectivity model of \citet{neuro,metric} (see Section \ref{neuroker}). We shall refer to this architecture as KerCNN. In the following, we first outline the proposed network architecture; we then introduce a testing framework to analyze the performance of the networks in three tasks of classification of corrupted images, focusing on types of image degradation where mechanisms of perceptual completion and global object analysis are required for correct classification.
 
 \subsection{Network architecture}\label{archi}
 
 The idea of the KerCNN architecture is to transpose the notion of connectivity of \citet{neuro,metric}, and notably the propagation of Eq. (\ref{imevol}), into the structure of a CNN. The lateral kernel $K_l$ associated to the $l$-th convolutional layer is defined as follows. First, a \emph{correlation kernel} $\tilde{K_l}$ is computed by taking the $L^2$ scalar product between the filters $\psi^l$:
 \begin{equation}\label{initK}
  \tilde{K_l}(i,j,f,g) = \nu\Big(\sum_{i',j',c} \psi^l_f(i',j',c)\: \cdot \: \psi^l_g(i-i',j-j',c)\Big),
 \end{equation}
 where $\nu$ is the sigmoidal activation function
 \[\nu(z) = \frac{1}{1+e^{-z}}.\]
 The indices in the sum are let vary as long as the product $\psi^l_f(i',j',c)\: \cdot \: \psi^l_g(i-i',j-j',c)$ does not vanish. Therefore, if the size of the bank of filters $\psi^l$ is $d_l \times d_l \times n_l \times n_{l-1}$, then the size of the kernel is obtained as $(2d_l-1)\times(2d_l-1)\times n_l \times n_l$.
 The final kernel is then obtained as
 \[K_l(i,j,f,g) = N[\tilde{K_l}](i,j,f,g),\]
 where $N$ is the same normalization operator introduced by \citet{coiflaf} and appearing in \citet{neuro}, see Eq. (\ref{laf}). Specifically, in the current case of a discrete, translation-invariant kernel $\mathcal{K}\big( (i,j,f),(0,0,g) \big) = \tilde{K_l}(i,j,f,g)$, the operator reads:
 \[N[\tilde{K_l}](i,j,f,g) := \frac{\tilde{K_l}^{(1)}(i,j,f,g)}{\sum_{i',j',g'} \tilde{K_l}^{(1)}(i',j',f,g')},\]
  where
  \[\tilde{K_l}^{(1)}(i,j,f,g) = \frac{\tilde{K_l}(i,j,f,g)}{\sum_{i',j',f'} \tilde{K_l}(i',j',f',g) \sum_{i',j',g'} \tilde{K_l}(i',j',f,g')}.\]
  \vskip 5 pt
  The update rule of a KerCNN layer is inspired by the iterative procedure outlined in the preceding section, designed to model the propagation of neural activity in V1:
  \begin{equation}\label{kcnnrule} \begin{cases}h_l^1 = s(\psi^l \ast h_{l-1}^{T_{l-1}} \:+\: b_l)\\
 h_l^t = \frac{1}{2}\Big(K_l \ast h_l^{t-1} \:+\: h_l^{t-1} \Big) \quad \text{for }\; 1 < t \leq T_l. \end{cases}\end{equation}
 The output of the $(l-1)$-th layer is first mapped to the $l$-th feature space through a feedforward step, yielding an activation $h_l^1$, which is then updated through convolution with the kernel $K_l$, as in (\ref{imevol}). The new output $h_l^2$ is defined by averaging between this updated activation $K_l \ast h_l^1$ and the original activation $h_l^1$. Note again an analogy with Eq. (\ref{meanfield}). The same procedure is repeated, yielding a sequence of activations $h_l^t$, until a fixed stopping time $T_l$ is reached. Note that each layer has its own stopping time: this yields a different KerCNN architecture for each combination of the stopping times $(T_1,\ldots,T_L)$ of the layers. If all stopping times are 1, the model coincides with the base CNN. We remark that convolutions with the kernel $K_l$ are taken with appropriate zero padding, so that the size of $h_l^{t}$ is preserved at every iteration. \\
 The intuitive idea here is that $K_l$ behaves like a ``transition kernel'' on the feature space of the $l$-th layer, slightly modifying its output according to the correlation between its filters: the activation of a filter encourages the activation of other filters highly correlated with it.

 \subsection{Task: stability to corrupted images}\label{task}
 
 We will show that the insertion of such structured lateral connections improves the performance of a CNN in tasks related to perceptual mechanisms of global shape analysis and integration. In particular, we focus on classification of \emph{corrupted} images. Given a labeled image dataset, each model is trained in a supervised way to perform classification. No corruption is applied to the images during the training phase. The actual experiment consists in analyzing the ability of the models to \emph{generalize} the classification to the degraded images, by comparing their classification accuracy on \emph{corrupted testing images}. We examine the following different kinds of image corruption.
 \begin{enumerate}
  \item Gaussian patches occluding the image, similar to the ones in \citet{tang}.
  \item Disruption of local contours, in analogy with the study presented by \citet{baker}, obtained by subdividing the image into horizontal or vertical strips and by shifting each of these strips by a random number of pixels $d \in \{0,\ldots,D\}$.
  \item Adversarial attacks through the Fast Gradient Sign Method (FGSM) of \citet{fgsm}. FGSM, one of the most popular attack methods, simply adjusts the input image by taking a \emph{gradient ascent} step to \emph{maximize} the loss function. Precisely, the perturbed image $I'$ is obtained as
 \begin{equation}\label{attack} I' = I + \varepsilon \cdot sign\big( \nabla_I \mathcal{L}(y(I),T) \big)\end{equation}
 where $\mathcal{L}$ is the loss function, as in Eq. (\ref{loss}).
 \end{enumerate}
 In all three cases, the amount of degradation can be quantified by one parameter: the standard deviation of the Gaussian patches, the maximum displacement $D$ of the strips, and the step $\varepsilon$ of the FGSM. 
 The more stable a model is to these perturbations, the slower the drop in performance w.r.t. the degradation parameter.

 \section{Results}\label{results}
 In this section, we first provide a complete analysis of the results obtained on the MNIST dataset \citep{mnist}: we compare the performance of a 2-layer CNN model with the ones of the corresponding KerCNN and RecCNN models, for varying stopping times $T_1$ and $T_2$ and for different types and amounts of image degradation (as outlined in Section \ref{task}). We then give a more synthetic report on the same study carried out on the Kuzushiji-MNIST \citep{kmnist}, Fashion-MNIST \citep{fashion} and CIFAR-10 \citep{cifar} datasets. All the experiments were implemented using PyTorch \citep{pytorch}.

 \subsection{MNIST}\label{resmnist}
\begin{figure}[htbp!]
 \centering
 \includegraphics[width=.7\textwidth]{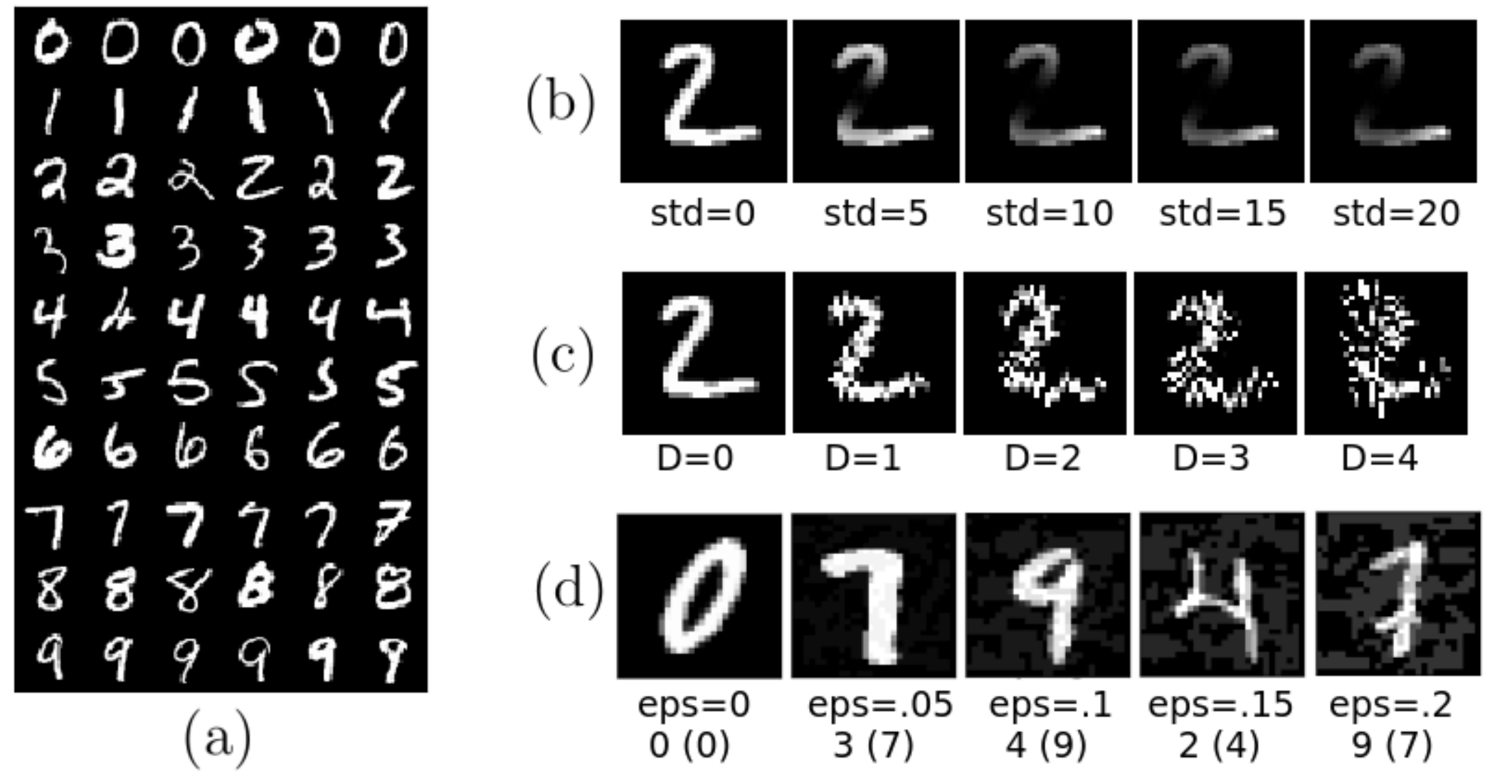}
 \caption{(a) A sample from the MNIST dataset. (b) A testing image corrupted by a Gaussian patch of increasing standard deviation. (c) A testing image corrupted by an increasing amount of local contour disruption $D$. (d) Testing images perturbed by applying FGSM to the base CNN, with increasing values of $\varepsilon$. Below each image, we display the classified label, as well as the correct label (in brackets). Apart from the unperturbed one ($\varepsilon=0$), all the images are misclassified by the CNN.}\label{MNIST}
 \end{figure}

 We start by considering the MNIST dataset \citep{mnist}, consisting of 70000 labeled $28\times 28$ grayscale images of handwritten digits from 0 to 9: see the sample in Figure \ref{MNIST}a. The default train-test split is 60000/10000. We retained a part of the images from the training set for validation-based early stopping, so that the final dataset used consisted of 50000 training samples, 10000 validation samples and 10000 testing samples. We trained the networks on the original training images, and we tested them on \emph{corrupted} testing images, according to the three types of degradation mentioned above. Some examples are displayed in Figure \ref{MNIST}b-d. 

 \begin{figure}[htbp!]
 \centering
 \includegraphics[width=.7\textwidth]{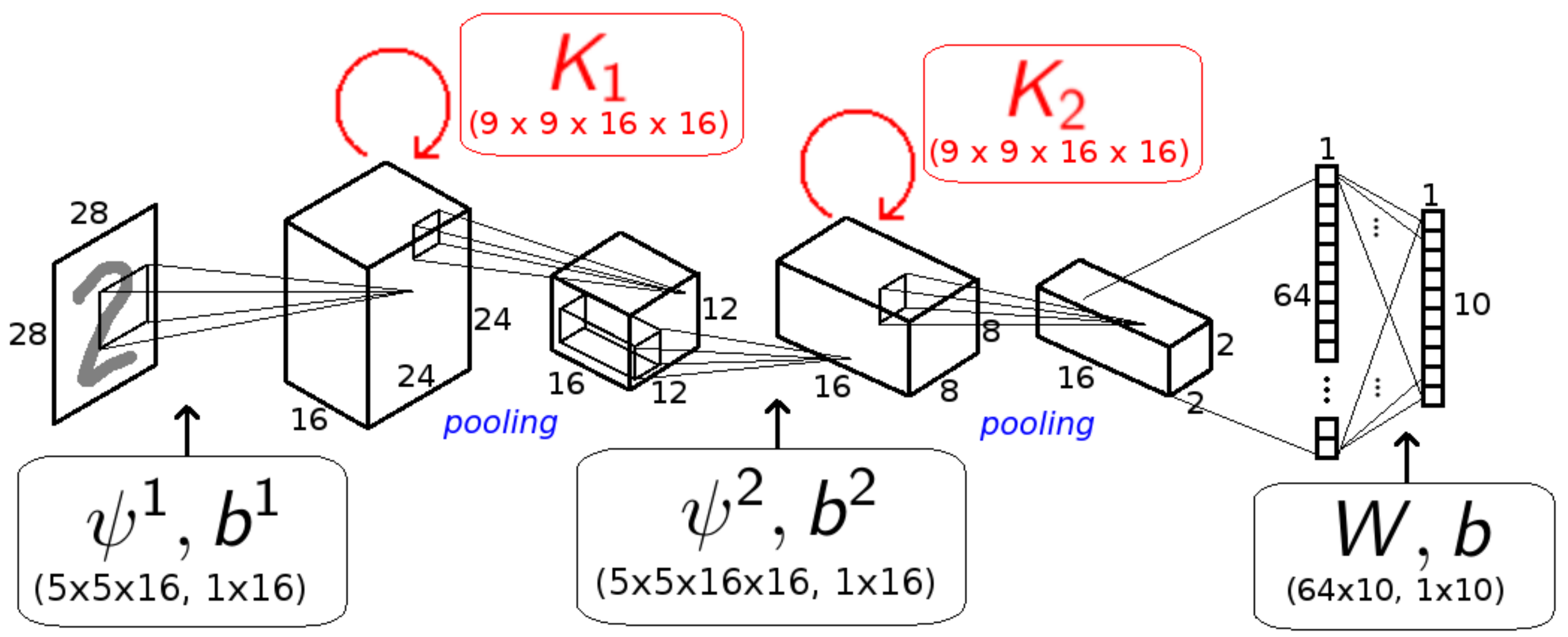}
 \caption{Our KerCNN model with structured lateral connections defined by kernels $K_1$ and $K_2$.}\label{model}
 \end{figure}
 \subsubsection{Base model}
 Our base model is a CNN with 2 hidden layers. We take 16 filters of size $5\times 5$ in the first convolutional layer and 16 filters of size $5\times 5 \times 16$ in the second convolutional layer, each followed by ReLU activation and max pooling, and a fully connected last layer followed by softmax activation. The total number of trainable parameters is 7482. We then compare this model with the one obtained from it by inserting the structured lateral connections. See Figure \ref{model} for a description of the model. The lateral kernels in this case have size $9 \times 9 \times 16 \times 16$. We also analyze the performance of the model obtained from the CNN by inserting recurrent connections according to the RecCNN model, i.e. through the update rule (\ref{rcnnrule}). As said before, lateral connections given by the kernels $K_l$ do not introduce new parameters in the starting CNN. On the other hand, the insertion of \emph{learned} lateral connections results in a model with more parameters than the base CNN: for example, the introduction of learned kernels of size $4\times 4 \times 16 \times 16$ in the first layer of the base model would add 4096 new parameters to the original 7482. In the following, we consider a 7482-parameter version of the RecCNN, obtained by decreasing the size of feedforward filters in order to compensate for the extra recurrent parameters, as in \citet{spoerer}.

 \subsubsection{Training details} All the models were trained with validation-based early stopping, for a maximum of 150 epochs. Adam optimizer was employed with the standard parameters indicated in \citet{adam}, a batch size of 50 and the Xavier initialization scheme \citep{xavier}; $L^2$ regularization with $\lambda=.0005$ was used. In the models including lateral connections (of any kind), recurrent dropout \citep{recdrop} with .2 probability was applied to the ``horizontal'' contributions. In RecCNNs, local response normalization (LRN) was applied after recurrent convolutional layers as in \citet{liang} and \citet{spoerer}. The training and testing images were z-score normalized according to the mean and standard deviation computed across the whole training set. For each architecture (i.e. each combination of stopping times $T_1$ and $T_2$), 10 nets initialized with different random seeds were trained. The results displayed in the following are obtained by testing all 10 nets and averaging the classification accuracy over trials. Note that the testing itself introduces a further element of randomness over trials, since the perturbations are applied to the images at each evaluation, yielding possibly different results. Error bars (95\% confidence intervals) are shown in the plots to keep track of the variability across initialization seeds and image perturbations.
 
 \subsubsection{Gaussian patches} We first consider testing images corrupted by \emph{occlusions} in the form of Gaussian ``bubbles'' at random locations over the image, similar to the ones considered by \citet{tang}. 
 Specifically, the image $I'$ obtained by modifying the original input $I$ through a patch centered at $(\overline{u}_1,\overline{u}_2)$ was implemented as:
  \[I'(u)=(I(u)-b)\cdot (1-g(u)) + b,\]
  where $g(u) := \frac{1}{2\pi\gamma^2}\exp\left(\frac{(u_1-\overline{u}_1)^2+(u_2-\overline{u}_2)^2}{2\gamma^2}\right)$ and $b$ is the ``background color'', chosen to be the value at the upper left angle of each image. See Figure \ref{MNIST}b. The number of patches per image was kept fixed to 4. In the following, we show the results of comparing the classification accuracy of the CNN and KerCNN models for varying amounts of image degradation (i.e. standard deviation $\gamma$ of the Gaussian bubbles, expressed in pixels) and for different stopping times of KerCNN.\\
 \begin{figure}[htbp!]
 \centering
 {\renewcommand{\baselinestretch}{0}
 \subfloat[\label{bubbles:T1}]{\includegraphics[width=.65\textwidth]{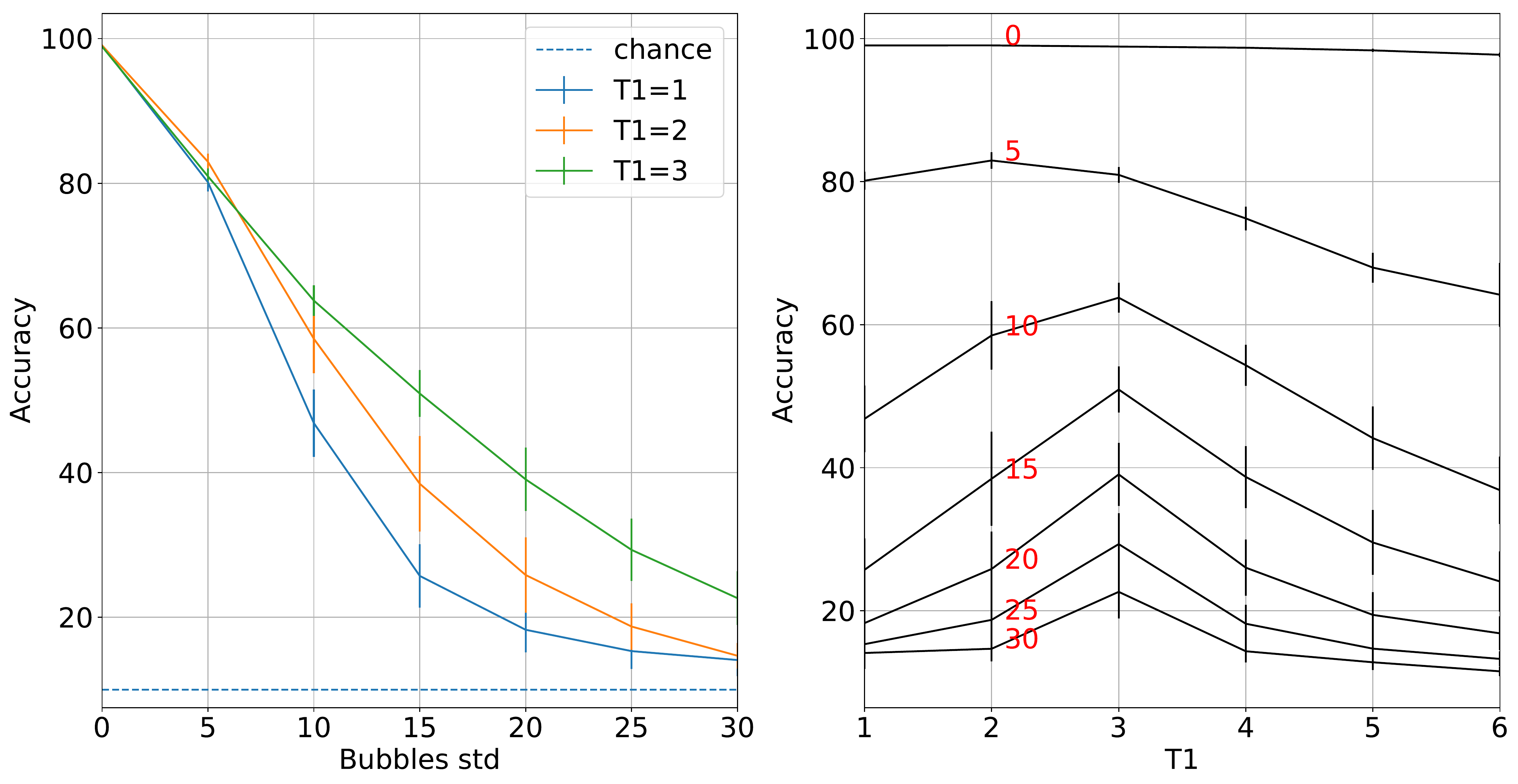}}\\
 \subfloat[\label{bubbles:T2}]{\includegraphics[width=.65\textwidth]{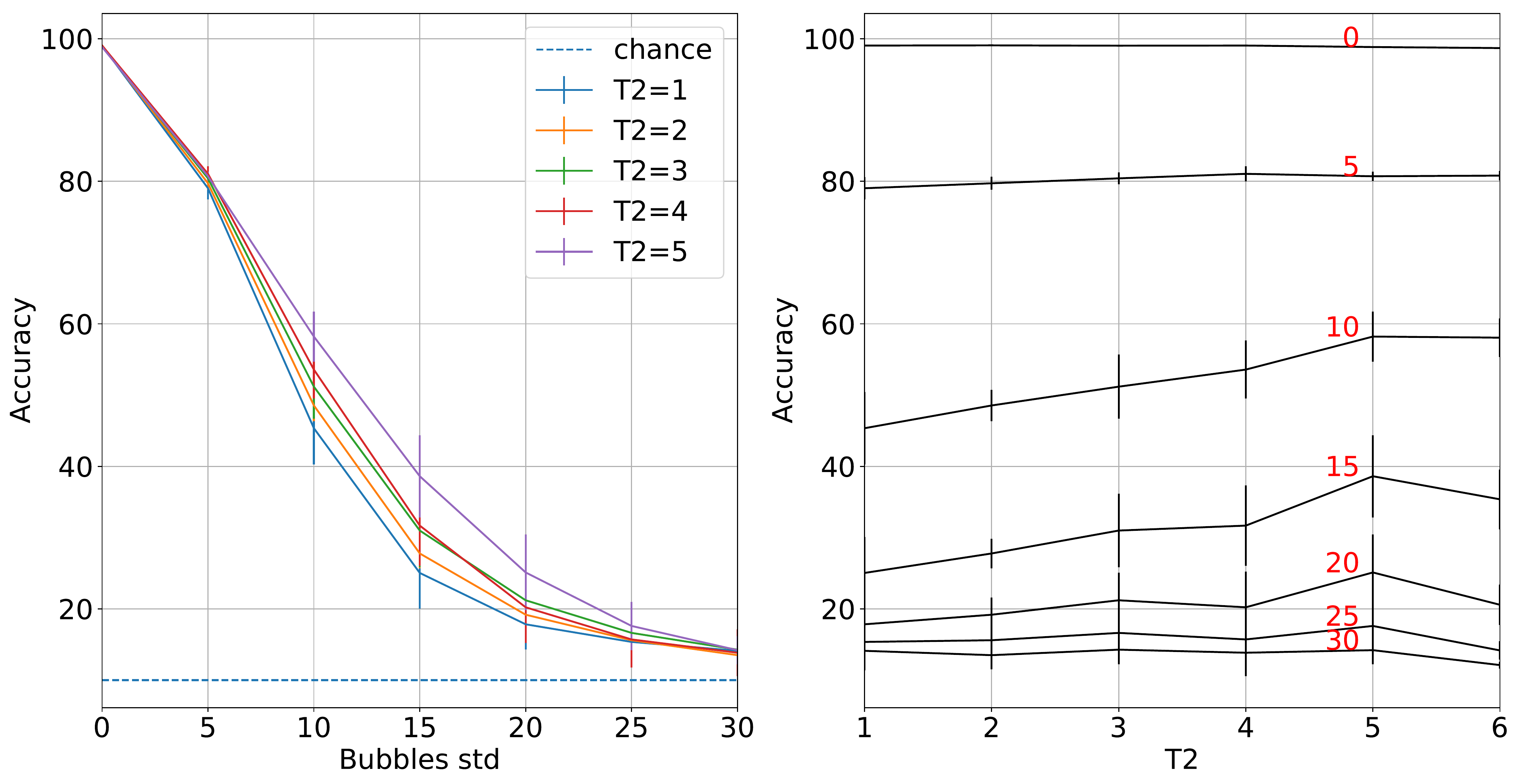}}
 }
 \caption{Results for MNIST testing images corrupted through Gaussian patches, for KerCNN with lateral connections in the first (A), resp. second layer (B). Left plots: accuracy ($y$-axis) at increasing values of $\gamma$ ($x$-axis), for stopping time $T_1=1,2,3$ (A), resp. $T_2=1,\ldots,5$ (B). Right plots: accuracy ($y$-axis) for increasing values ($x$-axis) of $T_1$ (A), resp. $T_2$ (B), for different values of degradation. Each curve refers to a value of $\gamma$, specified in red in correspondence of the curve.}\label{bubbles}
 \end{figure}
 
 We first examine the KerCNN defined by inserting lateral connections in the first layer of the base CNN. Figure \ref{bubbles:T1}(left) shows its classification accuracy for varying values of standard deviation $\gamma$ of the Gaussian patches. The three graphs displayed refer to different stopping times $T_1=1,2,3$. The chance level accuracy (10\%) is displayed as well (dashed blue line). For $T_1=1$, the model is the standard CNN with no lateral connections. The mean performance of these three nets on the original testing set ($\gamma=0$) is almost identical ($99.0\pm0.1 \%$). On the other hand, for increasingly degraded images the performance drops dramatically for the CNN ($T_1=1$, blue curve), while decaying much more slowly for increasing values of $T_1$. Note that the difference in classification accuracy between the CNN and the best KerCNN reaches $\sim25$ points.
 After reaching its optimal value ($T_1=2$ for $\gamma\leq5$ and $T_1=3$ for greater values), the performance drops again by taking further steps. For the sake of legibility, we displayed in the left plot only the curves up to the optimal value of $T_1$. The behavior of classification accuracy w.r.t. $T_1 \in \{1,\ldots,6\}$ can be best appreciated in the right plot of Figure \ref{bubbles:T1}, displaying a curve for each value of the standard deviation $\gamma$: for every $\gamma$, the accuracy increases w.r.t. $T_1$ until a maximum is reached, and then decreases again.\\
 
 We now analyze the performance of the KerCNN models with lateral connections:
  \begin{itemize}
  \item only in the second layer;
  \item in both layers.
 \end{itemize}
 Analogous to the preceding case, the optimal stopping time for the net with lateral connections in the second layer is $T_2=2$ for the original images, $T_2=4$ for a small degradation ($\gamma=5$) and $T_2=5$ for greater values of standard deviation. Figure \ref{bubbles:T2}(left) plots the accuracy against the level of degradation: we display the curves for $T_2=1,\ldots,5$; the accuracy w.r.t. stopping times $T_2 \in \{1,\ldots,6\}$ is plotted in Figure \ref{bubbles:T2}(right), where each curve corresponds to a level of image degradation. The results show the same pattern as before, although with a smaller improvement (up to $\sim15$ points between the base CNN and the model with optimal $T_2$).\\
 It is interesting to note that the optimal number of iterations shifts towards higher values (for both layers) as the size of the occlusions increases. As mentioned before, the kernel $K_l$ can be thought of as an \emph{anisotropic} transition kernel on the space of activations of the $l$-th layer. As such, the repeated application of the lateral contribution given by these kernels may be interpreted as a spreading of activation, around each spatial location, along those orientations that are most activated at that point. 
 \begin{figure}[htbp!]
  \centering
 \includegraphics[width=\textwidth]{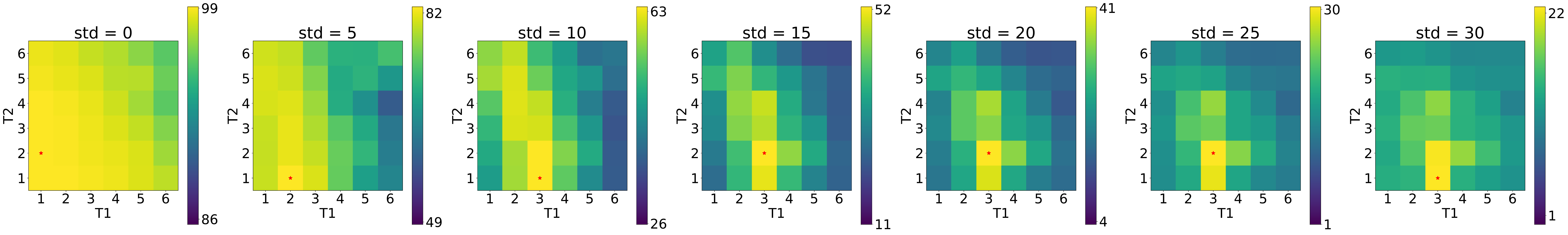}
 \caption{Classification accuracy (color-coded) for KerCNN for all combinations of $T_1, T_2 \in \{1,\ldots,6\}$, displayed for $\gamma=0,\ldots,30$. The maximum value of accuracy is marked by a red star onto the corresponding cell.}\label{comb_bubbles}
 \end{figure}
 Intuitively, this ``compensates'' for the gaps in the activation caused by the occlusions: the wider the gap, the higher the number of iterations of the kernel needed for the image to be consistently completed.\\
 We finally study the combinatorics of stopping times $T_1,T_2 \in \{1,\ldots,6\}$ in the two layers: Figure \ref{comb_bubbles} displays the results for different levels of image degradation. For each combination of $T_1$ ($x$-axis) and $T_2$ ($y$-axis), the mean accuracy over all trials (color-coded) is displayed. Note that the highest values of accuracy lie on a diagonal that shifts towards higher values of both $T_1$ and $T_2$ as the level of degradation increases. It is interesting to observe that, for $\gamma=15,20,25$, the optimal couple $(T_1,T_2)$, highlighted by a red star, is one involving lateral connections in both layers.

 \subsubsection{Local contour disruption} 
 In \citet{baker}, evidence is provided that the feature extraction performed by deep CNNs mostly relies on local edge relations, rather than on global object shapes. Their experiments showed that, conversely to human vision, the networks' performance was much more robust to global shape changes preserving local features, than to a disruption of local contours preserving the global information. We hypothesized that the insertion of structured lateral connections in CNNs could make the models more robust to these local perturbations.\\ 
 \begin{figure}[htbp!]
 \centering
 {\renewcommand{\baselinestretch}{0}
 \subfloat[\label{local:T1}]{\includegraphics[width=.65\textwidth]{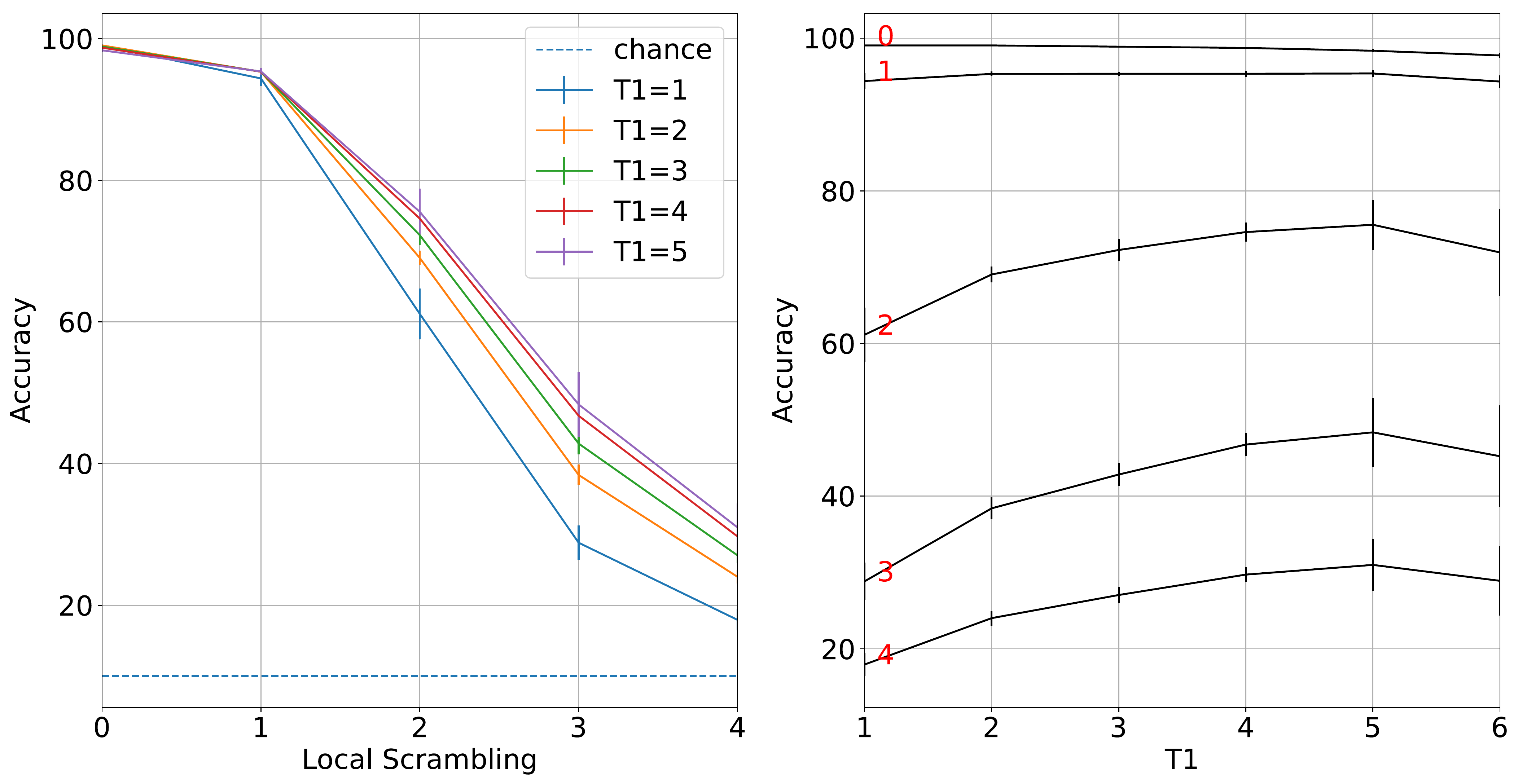}}\;
 \subfloat[\label{local:T2}]{\includegraphics[width=.65\textwidth]{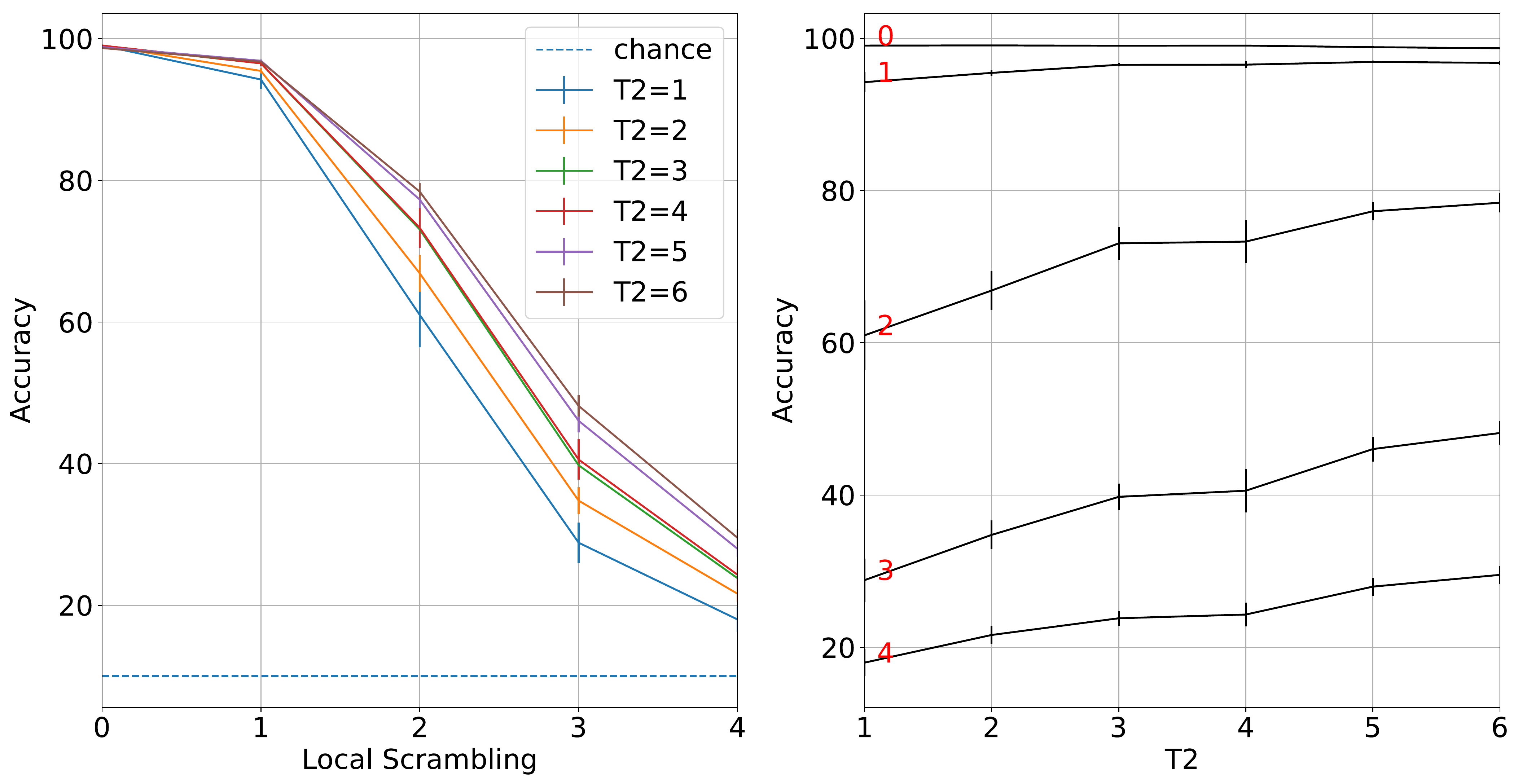}}
 }
 \caption{Results for MNIST testing images corrupted through local contour disruption, for KerCNN with lateral connections in the first (A), resp. second layer (B). Left plots: accuracy at increasing values of displacement $D$, for stopping time $T_1=1,\ldots,5$ (A), resp. $T_2=1,\ldots,5$ (B). Right plots: accuracy for increasing values of $T_1$ (A), resp. $T_2$ (B), for different values of degradation. Each curve refers to a value of $D$, displayed in red in correspondence of the curve.}\label{local}
 \end{figure}
 To automatically create a ``local scrambling'' of pixel information, we subdivided the images into horizontal strips and shifted each of these strips by a number of pixels $d$, randomly picked in $\{0,\ldots,D\}$; we then repeated the procedure by subdividing the modified image into vertical strips and by shifting them as well. For a small displacement ($D=1$), this produces a local degradation analogous to the one considered by \citet{baker}, where the local contours are corrupted but the connected components are preserved. For increasing values of $D$, the image is more and more disrupted, yet still roughly preserving its global structure. See Figure \ref{MNIST}c. As before, we compare the classification accuracy of the models for an increasing amount of degradation, given in this case by the maximum displacement $D$, which was kept the same for both horizontal and vertical strips. 
 \begin{figure}[htbp!]
  \centering
 \includegraphics[width=.8\textwidth]{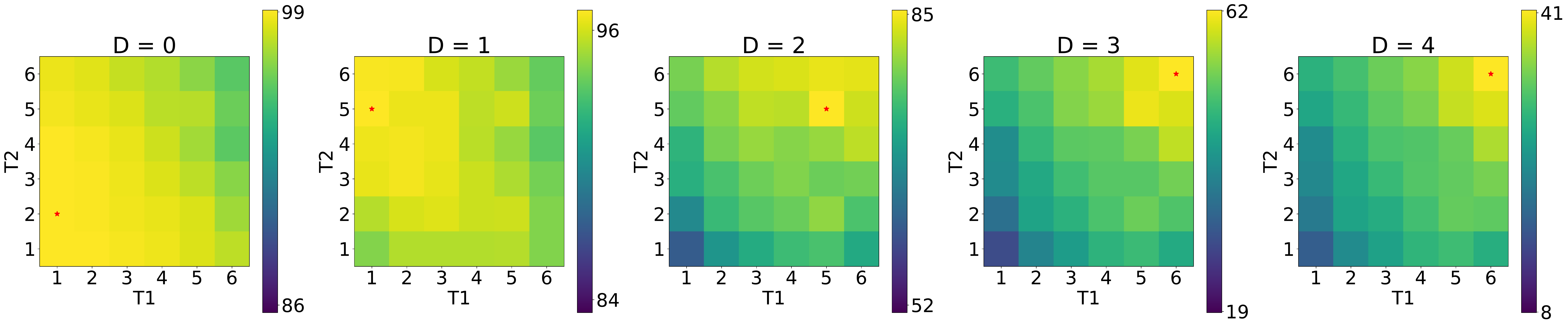}
 \caption{Classification accuracy (color-coded) for KerCNN for all combinations of $T_1, T_2 \in \{1,\ldots,6\}$, displayed for $D=0,\ldots,4$. The maximum value of accuracy is marked by a red star onto the corresponding cell.}\label{comb_local}
 \end{figure}
 In the present experiments, $D$ varies from 0 to 4 pixels. In this case, the performance of the models for $D\geq2$ turns out to rise for increasing stopping times up to $T_2=6$ for the models with lateral connections in the second layer, while there is a peak in performance at $T_1=5$ for the ones with lateral connections in the first layer: see Figure \ref{local}. A similar situation can be observed when analyzing the combinatorics of stopping times for the first and second layers, as shown in Figure \ref{comb_local}: the optimal couple of values $(T_1,T_2)$ shifts towards the maximum as the displacement $D$ increases, and the best accuracy is reached at $(T_1,T_2)=(6,6)$ above a certain amount of degradation.

 \subsubsection{Adversarial attacks}  
  Finally, we tested the robustness of our model to adversarial attacks via FGSM. Figure \ref{MNIST}d shows some examples of images obtained through (\ref{attack}) applied to the base CNN for MNIST, for increasing values of $\varepsilon$. 
  \begin{figure}[htbp!]
  \centering
 {\renewcommand{\baselinestretch}{0}
 \subfloat[\label{fgsm:T1}]{\includegraphics[width=.65\textwidth]{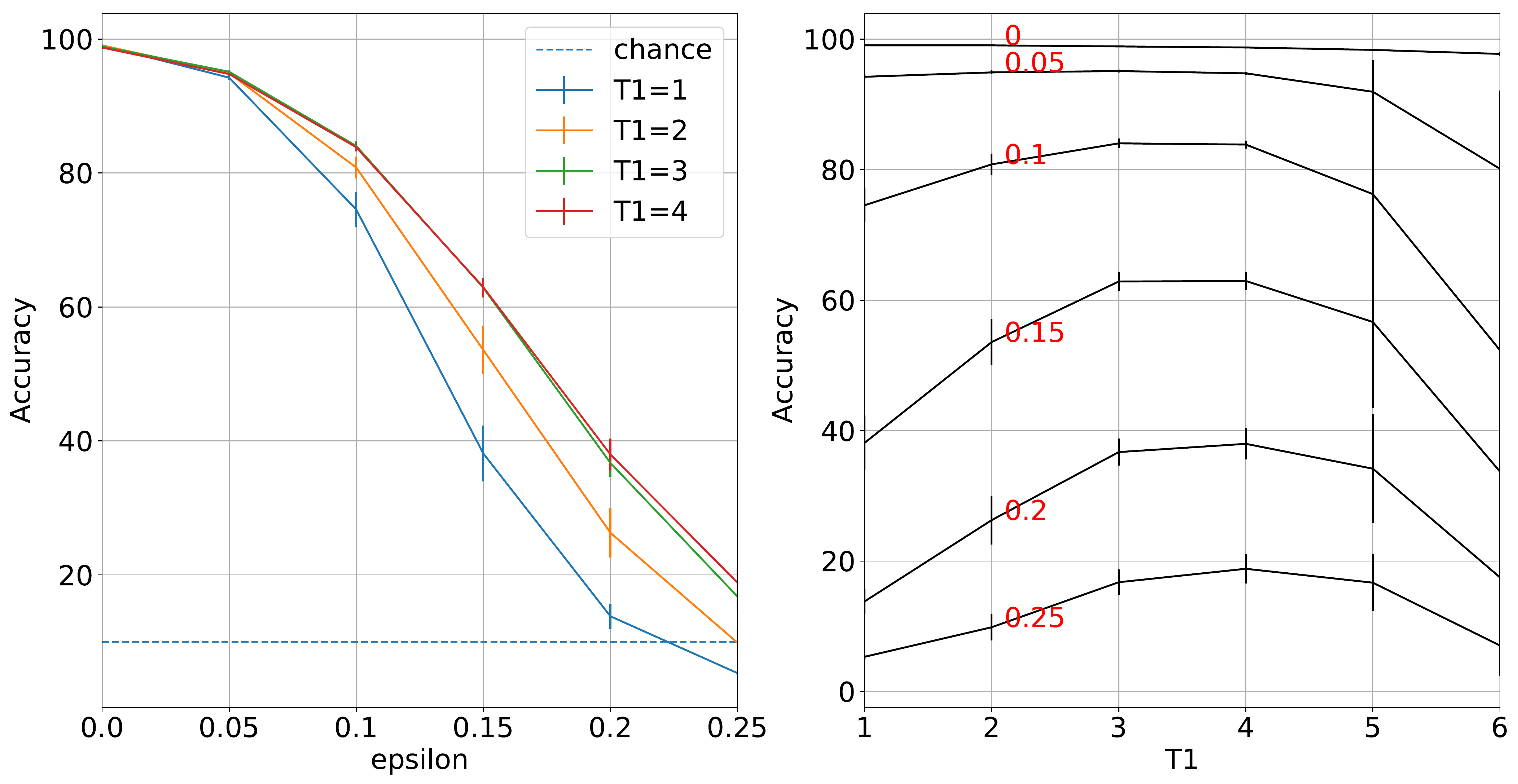}}\;
 \subfloat[\label{fgsm:T2}]{\includegraphics[width=.65\textwidth]{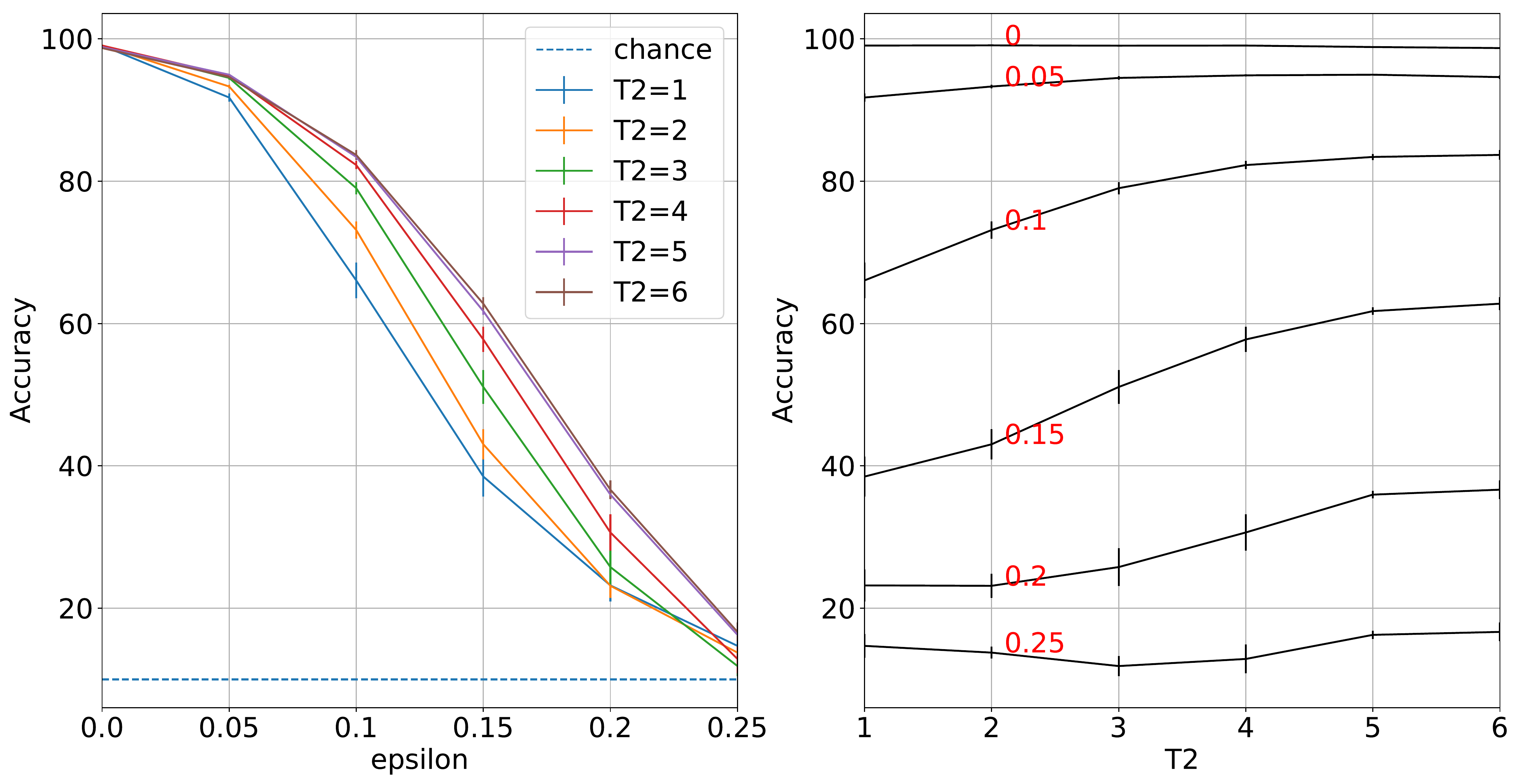}}
 }
 \caption{Results for MNIST testing images perturbed via FGSM, for KerCNN with lateral connections in the first (A), resp. second layer (B). Left plots: accuracy at increasing values of the FGSM parameter $\varepsilon$, for stopping time $T_1=1,\ldots,6$ (A), resp. $T_2=1,\ldots,6$ (B). Right plots: accuracy for increasing values of $T_1$ (A), resp. $T_2$ (B), for different values of degradation. Each curve refers to a value of $\varepsilon$, displayed in red in correspondence of the curve.}\label{fgsm}
 \end{figure}
 For sufficiently small $\varepsilon$, this perturbation results in an image that is almost identical to the original one to the human eye; however, these images are misclassified by the network.\\
  Again, we first examine the performance of the models with lateral connections in one layer at a time, for varying $T_1$ and $T_2$ respectively. Figure \ref{fgsm} displays the classification accuracies of these models for $T_1 \in \{1,\ldots,6\}$ and $T_2=1$ (A) and for $T_1=1$ and $T_2 \in \{1,\ldots,6\}$ (B). As before, the left figure plots the accuracy against the amount of degradation, with a curve for each stopping time $T_i$, while the right figure plots the accuracy against the stopping time $T_i$, with a curve for each value of $\varepsilon$. 
  \begin{figure}[htbp!]
  \centering
 \includegraphics[width=\textwidth]{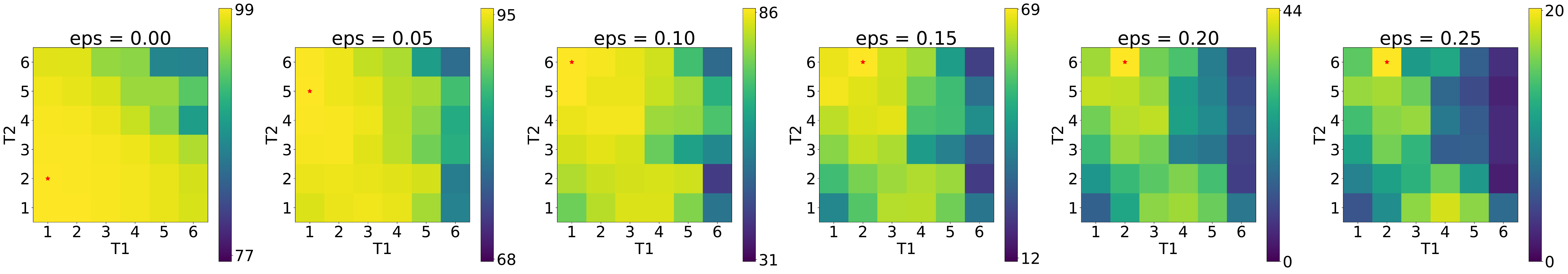}
 \caption{Classification accuracy (color-coded) for KerCNN for all combinations of $T_1, T_2 \in \{1,\ldots,6\}$, displayed for $\varepsilon=0,\ldots,0.25$. The maximum value of accuracy is marked by a red star onto the corresponding cell.}\label{comb_fgsm}
 \end{figure}
 Finally, Figure \ref{comb_fgsm} displays the analysis of the combinatorics of $T_1$ and $T_2$. Similarly to the case of Gaussian patches, the highest accuracy values lie on a diagonal. However, while in that case the optimal combination was clearly located around a single spot, two peaks develop in the current case, corresponding to either high values of $T_1$ and low values of $T_2$, or vice versa.\\

 We will summarize the main results obtained for all datasets in Table \ref{tab:overview}, showing the difference in mean percent accuracy between the base CNN and the optimal KerCNN model, along with the corresponding combination of stopping times $(T_1,T_2)$. A possible concern about our approach is the fact that we do not identify a combination that is optimal for all tasks, thus raising the issue of how to choose the stopping times when the amount of degradation is not known a priori. We nonetheless remark that, although the optimal combination of $(T_1,T_2)$ varies, the KerCNNs with $T_1, T_2 \in \{2,3\}$ outperform the base CNN in practically all the tasks.
 
 \FloatBarrier
 
 \subsubsection{Comparison with learned kernels}\label{cfr}
\begin{figure}[htbp!]
  \centering
 \subfloat[\label{rk:T1}]{\includegraphics[width=.4\textwidth]{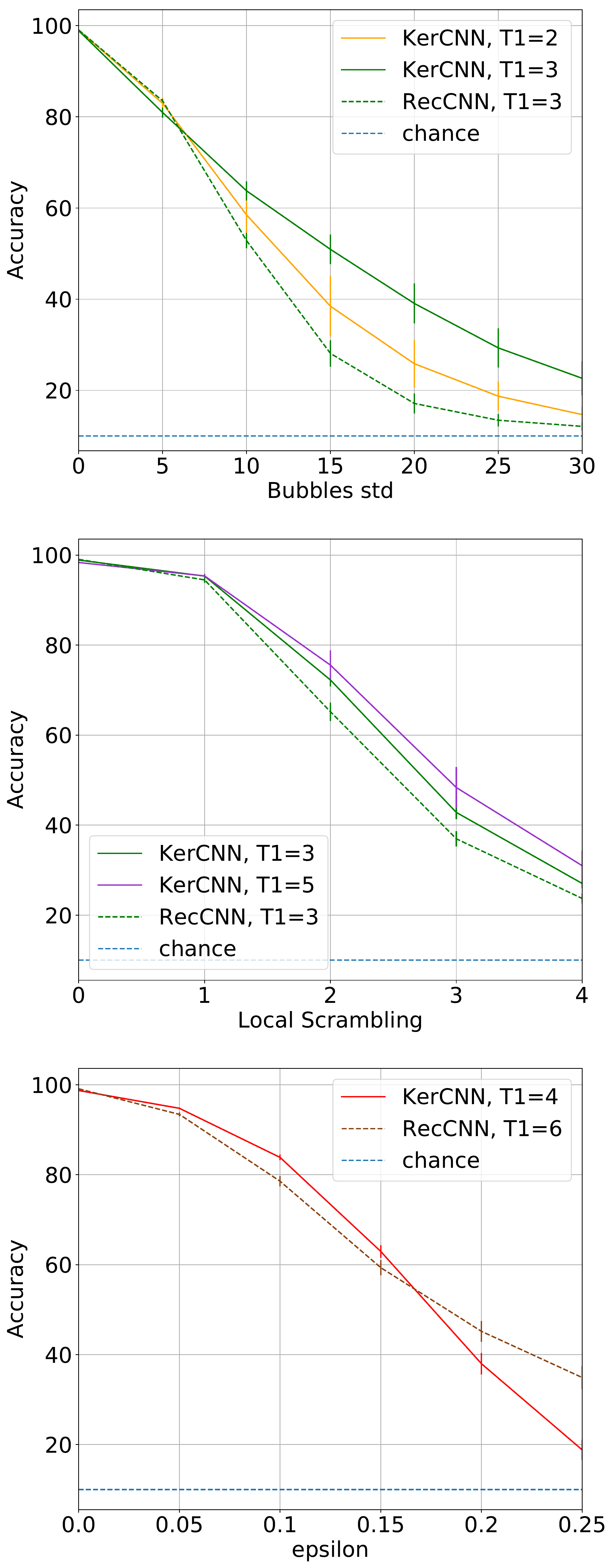}}\quad
 \subfloat[\label{rk:T2}]{\includegraphics[width=.4\textwidth]{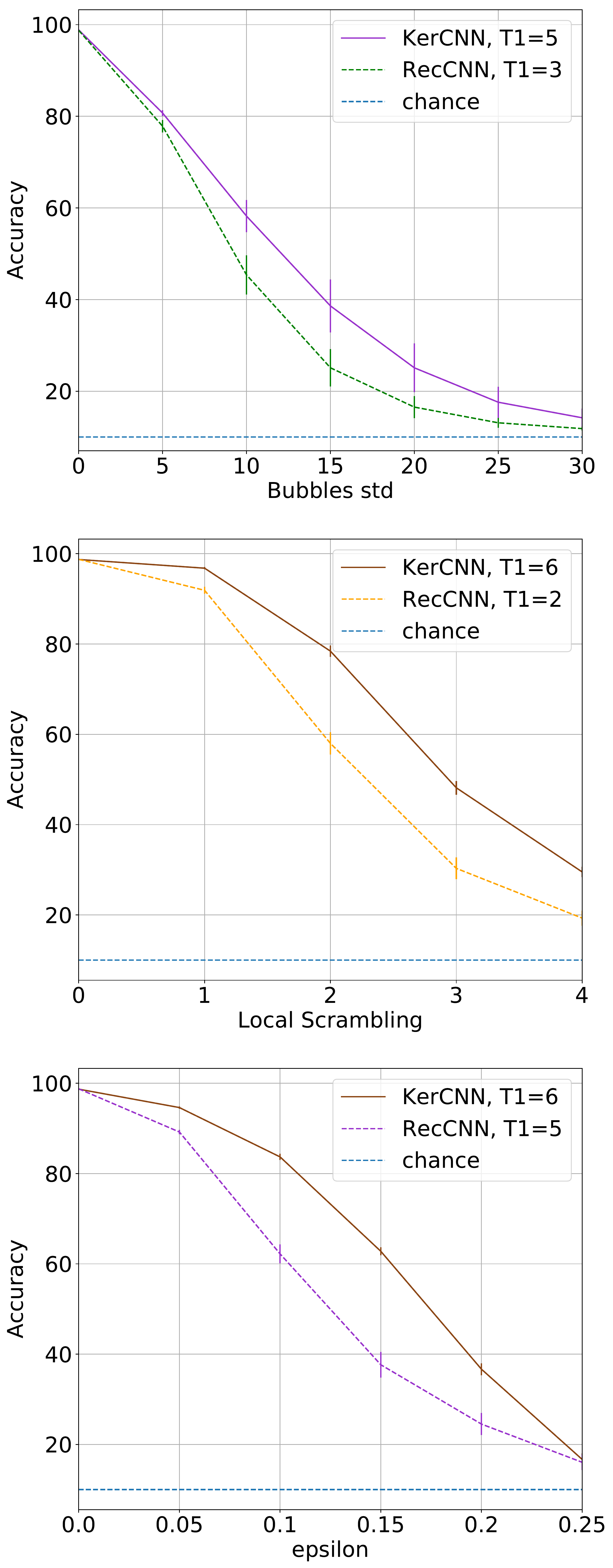}}
 \caption{Comparison between optimal KerCNN and optimal RecCNN. Top: Gaussian patches; middle: local edge disruption; bottom: adversarial attacks.
 }\label{rk}
 \end{figure}

 We now compare our model with the RecCNN architectures described above. Here, recurrent convolutional connections as described in Section \ref{reccnns}, with weights $\phi^1$ of size $4\times 4 \times 16 \times 16$, have been added in the first (resp. second) layer; the size of the feedforward weights of the second layer has been decreased to $3 \times 3 \times 16 \times 16$ to make the number of parameters match with the base CNN \citep[as in][]{spoerer}. The performance of these RecCNN models on the tasks examined before has been compared to the one of the base CNN, as well as with the corresponding KerCNNs. In most experiments, the RecCNN model did not reach better accuracies than the base CNN on corrupted images, although in some cases a pattern similar to the one seen for KerCNNs could be observed: in such cases, the performance increased until an optimal stopping time. However, the improvement in accuracy w.r.t. the CNN turned out to be much smaller than the one obtained by KerCNN models. Moreover, the geometric content of these learned lateral kernels is not evident and the iterative steps taken according to (\ref{rcnnrule}) do not seem to implement a kind of propagation -- a hint of this lies in the fact that the optimal stopping time for RecCNNs never depends on the amount of degradation of the testing images.\\
 In Figure \ref{rk}, we compare the accuracies of the KerCNN and RecCNN architectures for the corresponding optimal stopping times for each task. In all plots, the filled curves refer to KerCNN models, while the accuracy of RecCNNs is displayed by dashed curves. The color of each curve matches the one used for the corresponding stopping time in all the plots throughout the paper.\\
 Note that, in Figure \ref{rk:T1}(top), curves for KerCNN with both $T_1=2$ and $T_1=3$ are displayed. Although the KerCNN model with stopping time $T_1=3$ (orange curve) widely outperforms the optimal RecCNN for all values of standard deviation above 10, the RecCNN displays a higher accuracy with small occlusions. However, for these smaller patches the optimal stopping time for KerCNN is $T_1=2$ (green curve), and this model outperforms the best RecCNN for \emph{all values} of degradation. A similar situation can be observed in Figure \ref{rk:T1}(middle) for local edge disruption, where  both $T_1=3$ and $T_1=5$ curves are displayed for the KerCNN model.\\
 To sum up, the KerCNN model clearly outperforms the corresponding RecCNN architecture, when comparing the two for their respective best stopping times, for almost all tasks examined. It is interesting to note that the only case in which RecCNNs show a higher accuracy than KerCNNs for some values of degradation (only for lateral connections in the first layer) is when the images are perturbed via FGSM for $\varepsilon>0.2$. This suggests that, although the recurrent structure of RecCNNs may help improve the stability to ``noise-like'' perturbations, the absence of a geometric prior prevents them from implementing any mechanism of completion or contour integration. 
 It is worth noting that, in the study carried out by \citet{spoerer}, the networks were \emph{trained and tested} to recognize cluttered digits: in their experiments, RecCNNs significantly outperform the purely convolutional architectures, thus showing the benefits of recurrence in \emph{learning} challenging tasks. On the other hand, our study shows that this does not extend to the case where the networks are facing nuisances for which they were not specifically optimized. For such generalization task, our structured lateral connections inducing a geometric prior turn out to be much more effective.
 
 \subsection{Other datasets}
 
 In this last section, we provide a synthetic report of our results on some different datasets, namely Kuzushiji-MNIST \citep{kmnist}, Fashion-MNIST \citep{fashion} and CIFAR-10 \citep{cifar}. 
 We then illustrate our results through a summary table, which exhibits the improvement in accuracy obtained with the optimal $(T_1,T_2)$ w.r.t. the base CNN as an index of effectiveness of KerCNNs.
 
 \subsubsection{Kuzushiji-MNIST and Fashion-MNIST}
 
 In order to analyze the effect of our lateral connections on different images while keeping most of our settings unchanged, we examined two MNIST-like datasets: the Kuzushiji-MNIST dataset, containing 10 phonetic letters of hiragana, one of the components of the Japanese writing system; and the Fashion-MNIST dataset,  
 \begin{figure}[htbp!]
  \centering
 \includegraphics[width=\textwidth]{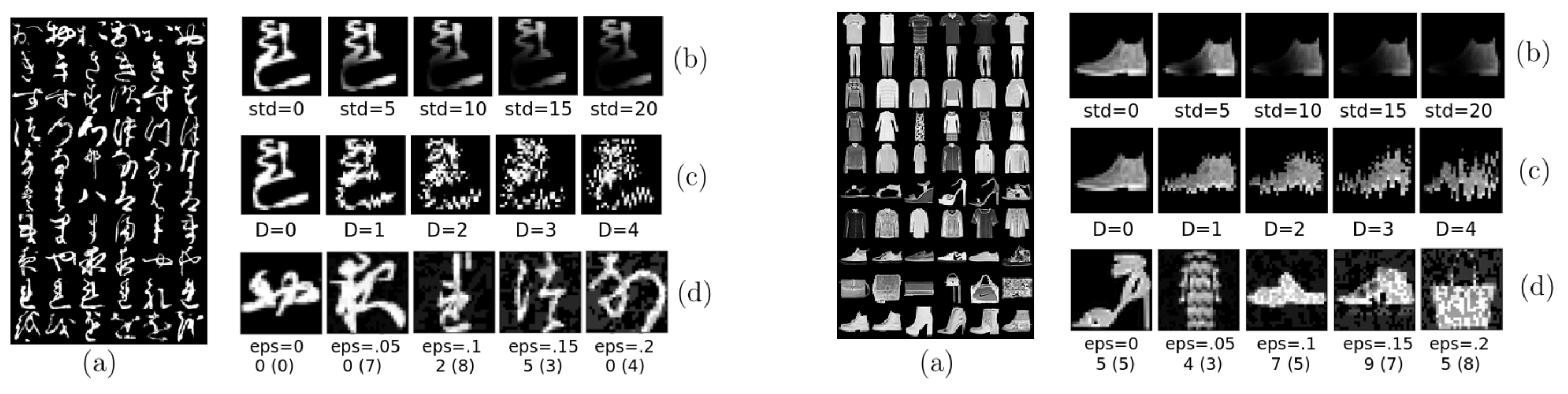}
 \caption{Examples from the Kuzushiji-MNIST (left) and Fashion-MNIST (right) datasets. For each database, the images display: (a) A sample from the dataset. Each row corresponds to a class. (b) A testing image corrupted by a Gaussian patch of increasing standard deviation. (c) A testing image corrupted by an increasing amount of local contour disruption $D$. (d) Testing images from different classes, perturbed by applying the FGSM to the base CNN with increasing values of $\varepsilon$. Below each image, we display the classified label, as well as the correct label (in brackets). Apart from the unperturbed image ($\varepsilon=0$), all the images are misclassified by the CNN.}\label{mnist_like}
 \end{figure} 
 consisting of Zalando's article images subdivided into 10 item categories (T-shirt/top, Trouser, Pullover, Dress, Coat, Sandal, Shirt, Sneaker, Bag, Ankle boot). Both datasets are made up of 70000 images of size 28$\times$28, with the same training-testing split as in MNIST. Figure \ref{mnist_like} displays, for each of these two datasets, some representatives of their 10 classes, as well as a few testing images corrupted by the three types of degradation examined. These have been implemented exactly as for MNIST, except for some changes in the range of degradation values considered (see Section \ref{resoverview}).\\
 Again, we considered a CNN with 2 hidden layers as a base model; the architecture is the same, except for the number of filters of the second layer which was set to 32 instead of 16, so that the total number of parameters becomes 14538. The training options were kept the same as before, except for the $L^2$ regularization parameter for Kuzushiji-MNIST which was set to $\lambda=.001$. With these choices, the mean accuracy of the base CNN is 93.13\% on Kuzushiji-MNIST and 89.86\% on Fashion-MNIST.

 \subsubsection{CIFAR-10}

\begin{figure}[ht!]
 \centering
 \includegraphics[width=.6\textwidth]{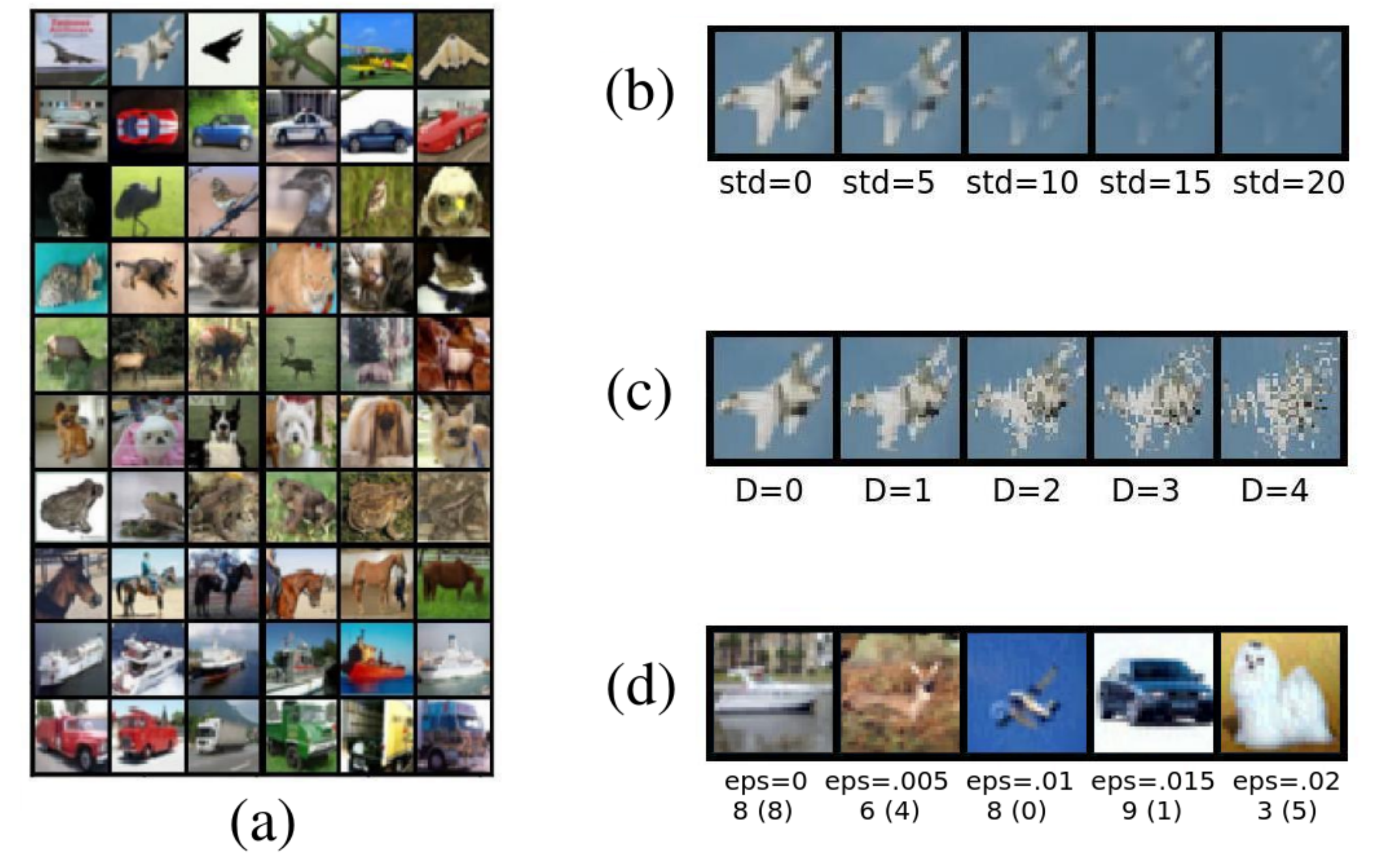}
 \caption{(a) A sample from the CIFAR-10 dataset. Each row corresponds to a class. (b) A testing image corrupted by a Gaussian patch of increasing standard deviation. (c) A testing image corrupted by an increasing amount of local contour disruption $D$. (d) Testing images from different classes, perturbed by applying the FGSM to the base CNN with increasing values of $\varepsilon$. Below each image, we display the classified label, as well as the correct label (in brackets). Apart from the unperturbed image ($\varepsilon=0$), all the images are misclassified by the CNN.}\label{cifar}
 \end{figure}

 The CIFAR-10 dataset consists of 60000 32$\times$32 color images in 10 classes (0:Airplane, 1:Automobile, 2:Bird, 3:Cat, 4:Deer, 5:Dog, 6:Frog, 7:Horse, 8:Ship, 9:Truck). In contrast with MNIST-like datasets, CIFAR-10 poses the significantly harder problem of recognizing objects in natural scene images. The dataset includes 50000 training images and 10000 test images. We extracted 10000 images from the training set to use for validation-based early stopping -- so that in our experiments the models were trained on 40000 samples, validated on 10000 samples and tested on 10000 samples. Figure \ref{cifar} shows some examples of (original as well as perturbed) testing images from CIFAR-10. The perturbations have been applied to the images by simply extending the former methods to three channels. Our base model is a 2-layer CNN with the same architecture as before, but with 64 and 128 filters respectively in the first and second convolutional layers. Moreover, since the images are RGB, the filters of the first layer have three channels in this case. The models were trained with early stopping for a maximum of 300 epochs.
 Stochastic gradient descent was employed with an initial learning rate of .01, which was automatically decreased by 1/10 when validation accuracy stopped increasing for 10 epochs. We used a batch size of 64 samples and an $L^2$ regularization parameter $\lambda=.001$. Also, dropout with .5 probability was employed in the last layer. The rest of the settings were kept the same as for the other datasets. Due to the longer training times, the results displayed for each architecture are obtained by averaging over 3 networks, instead of 10, trained with different random seeds. Moreover, we let vary the stopping times $T_i$ only in $\{1,2,3,4\}$. We remark that we are employing a rather small CNN (the total number of parameters is 214922), and no data augmentation is used. With these settings, the mean accuracy of the base CNN on CIFAR-10 is 75.64\%. We stress that our aim is to determine the \emph{improvement} brought by our lateral kernel: in order to better assess its effect, we thought it best to consider a simple network as a base model.

 \subsubsection{Results overview}\label{resoverview}
 
 Our results on all considered datasets are summarized in Table \ref{tab:overview}. For each dataset, the three row blocks correspond to the three types of perturbation examined. For each type of image degradation and each level of corruption, the table displays the mean percent accuracies of the base CNN and the best KerCNN, as well as their difference. The combination $(T_1,T_2)$ leading to the best KerCNN performance is shown next to the corresponding accuracy value. In the event when the optimal performance is reached by the CNN, the best $(T_1,T_2)\neq(1,1)$ is displayed.
 {\renewcommand{\arraystretch}{.7} 
 \begin{table}[htbp!]
 \centering
 \tiny
  \caption{Overview of the results for MNIST, Kuzushiji-MNIST, Fashion-MNIST and CIFAR-10. For each degradation value, the accuracy of the base CNN is compared to the one of the best KerCNN. The optimal $(T_1,T_2)\neq(1,1)$ is also shown for each case.}
 \label{tab:overview}
 \begin{tabular}{@{}lrrrrrrrr@{}}
 \toprule
      \multicolumn{8}{c}{\textbf{MNIST}}\vspace{5pt}\\
      \textbf{std $\gamma$} & \textbf{0} & \textbf{5} & \textbf{10} & \textbf{15} & \textbf{20} & \textbf{25} & \textbf{30} \\ 
      base CNN & 99.05\% & 80.13\% & 46.84\% & 25.73\% & 18.28\% & 15.33\% & 14.10\% \\ 
      best KerCNN & (1,2) 99.08\% & (2,1) 82.96\% & (3,1) 63.77\% & (3,2) 52.47\% & (3,2) 41.18\% & (3,2) 30.34\% & (3,1) 22.65\% \\
      difference & +0.04\% & +2.83\% & +16.94\% & \textbf{+26.74\%} & +22.90\% & +15.01\% & +8.55\% \vspace{5pt}\\  
      \textbf{Shift $D$} & \textbf{0} & \textbf{1} & \textbf{2} & \textbf{3} & \textbf{4} & & \\
    base CNN & 99.05\% & 94.39\% & 61.14\% & 28.83\% & 17.93\% & & \\ 
      best KerCNN & (1,2) 99.08\% & (1,5) 96.89\% & (5,5) 85.45\% & (6,6) 62.11\% & (6,6) 41.28\% & &  \\
    difference & +0.03\% & +2.51\% & +24.32\% & \textbf{+33.29\%} & +23.35\% & & \vspace{5pt}\\
    \textbf{FGSM $\varepsilon$} & \textbf{0} & \textbf{.05} & \textbf{.1} & \textbf{.15} & \textbf{.2} & \textbf{.25} & \\
    base CNN & 99.05\% & 94.23\% & 74.54\% & 38.12\% & 13.82\% & 14.83\% & \\ 
    best KerCNN & (1,2) 99.08\% & (1,5) 95.71\% & (1,6) 86.89\% & (2,6) 69.20\% & (2,6) 44.37\% & (2,6) 20.16\% & \\
    difference & +0.04\% & +1.47\% & +12.35\% & \textbf{+31.08\%} & +30.55\% & +5.33\% & \\
     \midrule
      \multicolumn{8}{c}{\textbf{Kuzushiji-MNIST}}\vspace{5pt}\\
      \textbf{std $\gamma$} & \textbf{0} & \textbf{5} & \textbf{10} & \textbf{15} & \textbf{20} & \textbf{25} & \textbf{30} \\
      base CNN & 93.13\% & 74.20\% & 39.67\% & 21.22\% & 14.81\% & 36.79\% & 30.38\% \\ 
      best KerCNN & (1,2) 93.13\% & (2,1) 75.72\% & (3,3) 59.96\% & (3,3) 51.44\% & (3,3) 43.69\% & (3,3) 12.39\% & (3,3) 11.41\% \\
      difference & +0.00\% & +1.53\% & +20.29\% & \textbf{+30.22\%} & +28.89\% & +24.40\% & +18.97\% \vspace{5pt}\\ 
     \textbf{Shift $D$} & \textbf{0} & \textbf{1} & \textbf{2} & \textbf{3} & \textbf{4} & & \\
   base CNN    & 93.13\% & 85.06\% & 61.62\% & 42.15\% & 31.49\% & & \\ 
    best KerCNN & (1,2) 93.13\% & (1,4) 87.91\% & (3,4) 73.60\% & (5,2) 59.15\% & (5,3) 47.48\% & & \\
    difference & +0.00\% & +2.85\% & +11.99\% & \textbf{+17.00\%} & +16.00\% & & \vspace{5pt}\\
    \textbf{FGSM $\varepsilon$} & \textbf{0} & \textbf{.05} & \textbf{.1} & \textbf{.15} & \textbf{.2} & \textbf{.25} & \\
   base CNN    & 93.13\% & 65.03\% & 28.15\% & 11.28\% & 6.36\% & 3.95\% & \\ 
    best KerCNN & (1,2) 93.13\% & (1,5) 74.08\% & (1,6) 48.91\% & (1,6) 25.63\% & (5,5) 13.74\% & (5,6) 7.76\% & \\
    difference & +0.00\% & +9.05\% & \textbf{+20.76\%} & +14.35\% & +7.38\% & +3.81\% & \\
    \midrule
      \multicolumn{8}{c}{\textbf{Fashion-MNIST}}\vspace{5pt}\\
    \textbf{std $\gamma$} & \textbf{0} & \textbf{5} & \textbf{10} & \textbf{15} & \textbf{20} & \textbf{25} & \textbf{30} \\
     base CNN    & 89.86\% & 72.03\% & 49.07\% & 32.47\% & 22.43\% & 17.13\% & 14.18\% \\ 
      best KerCNN & (1, 3) 90.02\% & (3, 1) 73.37\% & (3, 2) 55.44\% & (3, 2) 43.03\% & (3, 2) 31.71\% & (4, 4) 25.55\% & (4, 4) 22.57\% \\
      difference & +0.16\% & +1.33\% & +6.37\% & \textbf{+10.55\%} & +9.28\% & +8.42\% & +8.39\% \vspace{5pt}\\  
    \textbf{Shift $D$} & \textbf{0} & \textbf{1} & \textbf{2} & \textbf{3} & \textbf{4} & & \\
   base CNN    & 89.86\% & 77.27\% & 58.58\% & 44.33\% & 34.81\% &  & \\ 
      best KerCNN & (1, 3) 90.02\% & (4, 3) 83.69\% & (5, 4) 72.18\% & (6, 6) 66.43\% & (6, 6) 60.87\% &  & \\
    difference & +0.16\% & +6.42\% & +13.61\% & +22.10\% & \textbf{+26.06\%} & & \vspace{5pt}\\
  \textbf{FGSM $\varepsilon$} & \textbf{0} & \textbf{.02} & \textbf{.04} & \textbf{.06} & \textbf{.08} & \textbf{.1} & \\
   base CNN    & 89.86\% & 53.81\% & 31.49\% & 18.53\% & 13.01\% & 10.13\% &  \\ 
      best KerCNN & (1, 3) 90.02\% & (2, 6) 70.48\% & (2, 6) 54.83\% & (2, 6) 42.78\% & (2, 6) 32.84\% & (2, 6) 25.57\% &  \\
    difference & +0.16\% & +16.67\% & +23.34\% & \textbf{+24.25\%} & +19.82\% & +15.45\% & \\
      \midrule
      \multicolumn{8}{c}{\textbf{CIFAR-10}}\vspace{5pt}\\
     \textbf{std $\gamma$} & \textbf{0} & \textbf{5} & \textbf{10} & \textbf{15} & \textbf{20} & \textbf{25} & \textbf{30} \\
      base CNN  & 75.64\% & 58.22\% & 32.84\% & 22.89\% & 19.27\% & 17.97\% & 17.40\% \\ 
      best KerCNN  & (2, 1) 75.57\% & (2, 1) 58.08\% & (2, 1) 32.90\% & (2, 2) 23.57\% & (3, 2) 20.53\% & (3, 2) 19.33\% & (4, 1) 18.89\% \\
      difference & - 0.07\% & - 0.14\% & +0.06\% & +0.67\% & +1.26\% & +1.36\% & \textbf{+1.49\%} \vspace{5pt}\\ 
    \textbf{Shift $D$} & \textbf{0} & \textbf{1} & \textbf{2} & \textbf{3} & \textbf{4} & & \\
     base CNN     & 75.64\% & 41.7\% & 27.70\% & 23.71\% & 21.91\% & & \\ 
      best KerCNN  & (2, 1) 75.57\% & (4, 4)52.97\% & (4, 4) 43.33\% & (4, 4) 36.72\% & (4, 4) 31.99\% & & \\
    difference & - 0.07\% & +11.27\% & \textbf{+15.63\%} & +13.02\% & +10.08\% & & \vspace{5pt}\\
   \textbf{FGSM $\varepsilon$} & \textbf{0} & \textbf{.005} & \textbf{.01} & \textbf{.015} & \textbf{.02} & \textbf{.025} & \\
    base CNN    & 75.64\% & 42.9\% & 21.80\% & 10.83\% & 5.32\% & 2.86\% & \\ 
      best KerCNN  & (2, 1) 75.57\% & (2, 3) 51.25\% & (3, 4) 35.55\% & (4, 4) 25.58\% & (4, 4) 18.66\% & (4, 4) 13.53\% & \\
    difference & - 0.07\% & +8.35\% & +13.75\% & \textbf{+14.75\%} & +13.33\% & +10.67\% & \\
    \bottomrule
     \end{tabular}
\end{table}
}

 For what concerns Kuzushiji-MNIST, the best performance improvement for images occluded by Gaussian patches is comparable to the one obtained for MNIST. However, a greater contribution of the second layer's kernel can be observed: that is, the optimal combinations of stopping times display larger values of $T_2$ for this type of degradation. This may be due to the more frequent occurrence of complex patterns requiring a ``higher order'' analysis (such as crossings and loops) w.r.t. MNIST. On the other hand, on images subject to local displacement, the values of $T_1$ and $T_2$ bringing to the best accuracy are overall smaller, also leading to significantly smaller differences in performance relative to MNIST. In fact, the abundance of small details in such characters makes this kind of perturbation far more disruptive than it is for images like MNIST's digits: even a small displacement may completely destroy some tiny yet characterizing features. Finally, the results for adversarial attacks with small values of $\varepsilon$ are analogous to the ones obtained for digits, although with a faster decay in accuracy. On the other hand, although a configuration different from MNIST is observed for $\varepsilon\geq.2$, the accuracy values are around (or even below) chance level in these cases, which makes somewhat pointless to speculate about them.
 
 Let us now examine the results obtained for the Fashion-MNIST dataset. As for the images occluded by Gaussian patches, the slightly increased contribution of the second layer w.r.t. MNIST is again probably due to the heterogeneity of features characterizing these images, including both extended contours and tiny, intricate line patterns. For this type of perturbation, the improvement provided by our lateral connections is more moderate than it is for the preceding datasets, reaching a maximum accuracy difference of $\sim$10\%. This may depend upon such images being largely composed by ``solid color'' areas rather than lines. Intuitively, when an occlusion falls in the middle of one such area, it does not interrupt a curve or a contour: therefore, the activation values of filters sensitive to local orientation is very low at these locations and consequently the action of the kernel on them is less relevant. On the other hand, the perturbation obtained by shifting horizontal and vertical strips does not affect constant areas, while it consistently disrupts the image edges. Moreover, differently from Kuzushiji-MNIST's characters, global shapes rather than local details are markedly characterizing for discriminating between Fashion-MNIST classes. This makes our lateral connections particularly suited to manage this kind of perturbation. Indeed, a far greater improvement in the CNN performance can be observed w.r.t. Kuzushiji-MNIST in this case, especially for large values of the displacement $D$: as an example, for $D=4$, the $\sim$35\% accuracy obtained by the base CNN rises to $\sim$60\% with the optimal KerCNN model. Finally, for what concerns adversarial attacks, we considered values of $\varepsilon$ varying in a smaller range, since the decay in performance for this dataset turned out to be much faster; namely, we took $\varepsilon \in \{0,.02,.04,.06,.08,.1\}$. Again, up to this rescaling, the results are analogous to the other datasets.
 
 As for CIFAR-10, the performance of CNNs and KerCNNs on images corrupted by Gaussian patches is comparable for all values of $\gamma$, with a slight advantage for KerCNNs for occlusions large enough ($\gamma>5$). In our view, such ``insensitivity'' of lateral kernels to this type of perturbation may be linked to the increased difficulty of dealing with color images -- indeed, this aspect certainly requires further investigation. On the other hand, the improvement obtained by KerCNNs w.r.t. CNNs for images subject to edge disruption and adversarial attacks is still consistent (up to $\sim$15\%). Note that the value of $\varepsilon$ for adversarial attacks in this case was let vary in $\{0,.005,.01,.015,.02,.025\}$ (again due to the faster decay in accuracy w.r.t. $\varepsilon$). 
 
 Overall, we believe that the global results are very promising, both for what concerns the effectiveness of the model for image recognition under challenging conditions, and from the point of view of its interpretation linked to biological vision.

 \FloatBarrier
 
 \section{Conclusion}
  
 In this article we introduced KerCNN, a modification of a CNN architecture given by the addition of biologically inspired lateral connections. Such connections are determined by convolutional kernels iteratively applied to the output of each convolutional layer, and defined by a notion of correlation between the filters of that layer, as in the cortical connectivity model of \citet{neuro,metric}. This allows to establish a link between the geometry of feedforward and lateral connections, as the latter are defined in terms of the former. Moreover, since the lateral kernels are a deterministic function of the convolutional filters, the number of parameters of the original CNN is left unchanged -- thus allowing a fair comparison between a base CNN architecture and the KerCNNs obtained from it. 
  
 The models were compared on their ability to \emph{generalize} a learned image classification task to unseen corrupted inputs. The types of perturbation applied to the images were chosen to disrupt discriminative local information, so as to ``force'' the nets to perform an integration of \emph{context data} to correctly recognize the corrupted input. The biological reason for choosing this testing framework was the close bond between anatomical lateral connections and perceptual phenomena linked to global shape analysis. In fact, our study revealed that the insertion of the proposed lateral connections in a 2-layer CNN critically enhanced its stability to all types of perturbation examined. Moreover, such improvement was not observed when introducing \emph{learned} lateral kernels as in \citet{liang} and \citet{spoerer}. This suggests that the geometric information encoded in our lateral kernel has a meaningful role in implementing mechanisms of pattern completion and contour integration, to compensate for the missing information in the corrupted testing images. We remark that such mechanisms are ``spontaneous'' to the effect that that they are not enforced during the training stage: indeed, the networks were only trained to classify uncorrupted images.
 
 The main analysis was carried out on the MNIST dataset, and then extended to a few more image datasets. Notably, promising results were obtained on natural images from the CIFAR-10 dataset. As a future development, we intend to test our model on bigger images and on richer datasets. It would also be interesting to examine the connectivity kernels obtained for non-image data and for different tasks: as an example, the regularity enforced by our lateral kernels may be helpful for problems of sound source separation. 
 
 Another natural advancement would be to consider deeper architectures. Indeed, although the proposed architecture was motivated by a model for early visual areas, its flexibility could make it suitable for recovering patterns in higher level processing as well. An analysis of the different feature information encoded in the kernels associated to each layer may help gain better insight into the analysis carried out by the networks at each stage of their processing.
 
 \subsection*{Acknowledgments}
 The authors have been supported by Horizon 2020 Project ref. 777822: GHAIA.


\end{document}